\documentclass[UTF8]{article}

\usepackage[preprint]{neurips_data_2024}

\usepackage[utf8]{inputenc} 
\usepackage[T1]{fontenc}    
\usepackage{hyperref}       
\usepackage{url}            
\usepackage{booktabs}       
\usepackage{amsfonts}       
\usepackage{nicefrac}       
\usepackage{microtype}      
\usepackage{xcolor}         
\usepackage{CJKutf8}
\usepackage{caption}
\usepackage{graphicx, subfig}
\usepackage{tabularray}
\usepackage{longtable}
\usepackage{array}
\usepackage{booktabs}
\usepackage{xcolor}
\usepackage{geometry}
\usepackage{fancybox}
\usepackage{authblk}

\title{IPEval: A Bilingual Intellectual Property Agency Consultation Evaluation Benchmark for Large Language Models}

\author[1]{\textbf{Qiyao Wang}}
\author[2]{\textbf{Jianguo Huang}}
\author[1]{\textbf{Shule Lu}}

\author[3]{\\{\textbf{Yuan Lin*}}}
\author[2]{\textbf{Kan Xu}}
\author[2]{\textbf{Liang Yang}}
\author[2]{\textbf{Hongfei Lin}}

\affil[1]{School of Future Technology, Dalian University of Technology}
\affil[2]{School of Computer Science and Technology, Dalian University of Technology}
\affil[3]{School of Public Administration and Policy, Dalian University of Technology}

\affil[ ]{\texttt{\{wangqiyao, HuangJianguo, lsl2021\}@mail.dlut.edu.cn}}
\affil[ ]{\texttt{\{zhlin, xukan, liang, hflin\}@dlut.edu.cn}}
\affil[ ]{\url{https://ipeval.github.io/}}

\begin{document}

\maketitle
\begin{CJK}{UTF8}{gbsn}
\begin{abstract}
With the rapid development of Large Language Models (LLMs) in vertical domains, attempts have been made to the field of intellectual property (IP). However, there is currently no evaluation benchmark specifically for assessing the understanding, application, and reasoning abilities of LLMs in the IP domain. To address this issue, we introduce IPEval, the first capability evaluation benchmark designed for IP agency and consulting tasks. IPEval consists of 2657 multiple-choice questions, divided into four major capability dimensions: creation, application, protection, and management. These questions cover eight areas: patent rights which including inventions, utility models, and designs, trademarks, copyrights, trade secrets, integrated circuit layout design rights, geographical indications, and related laws. We designed three evaluation methods: zero-shot, 5-few-shot, and Chain of Thought (CoT) for seven kinds of LLMs with varying parameters, primarily using either English or Chinese. The study results indicate that the GPT series and Qwen series models demonstrate stronger performance in English tests, while Chinese-major LLMs, such as the Qwen series, outperform GPT-4 in Chinese tests. Specialized legal domain LLMs, such as the fuzi-mingcha and MoZi, still significantly lag behind general-purpose LLMs of comparable parameter sizes in IP performance. This highlights the necessity and substantial potential for developing more specialized LLMs with stronger IP abilities. We also analyze the models' capabilities in terms of the regional and temporal aspects of IP, emphasizing that IP domain LLMs need to clearly understand the differences in IP laws across different regions and their dynamic changes over time. We hope IPEval can provide an accurate assessment of LLM capabilities in the IP domain and encourage researchers interested in IP to develop LLMs with richer IP knowledge.
\end{abstract}

\section{Introduction}
\label{1}
Research has established evaluation benchmarks for large language models (LLMs) across diverse vertical domains, with the medical, legal, and educational sectors experiencing the most rapid development, driven by their extensive application potential. In the medical field, MedQA\cite{jin2021disease} is a prominent example, using USMLE-style open-domain multiple-choice questions to assess a model's ability to address medical issues and diagnose diseases. In the legal field, LawBench\cite{Fei2023LawBenchBL} stands out, precisely evaluating a model's legal capabilities across three cognitive levels: legal knowledge memory, comprehension, and application. In the educational field, E-EVAL\cite{hou2024eval} is representative, compiling exam questions from various stages and subjects of Chinese education to evaluate LLMs' grasp of basic educational knowledge and their reasoning ability to solve fundamental problems across different educational stages and disciplines.

However, we note that, there are few evaluation benchmarks for the Intellectual Property (IP) domain. HUPD\cite{suzgun2024harvard} compiles publicly available patent application data from the United States Patent and Trademark Office (USPTO), categorizing IP tasks into patent subject classification, language modeling, summarization, and other text analysis tasks. MoZIP\cite{ni2024mozip} focuses on LLMs' understanding of multilingual IP-related concepts and regulations, rather than evaluating LLMs' capabilities in understanding, applying, and reasoning with IP knowledge as an IP agent.

To address the lack of benchmark tests for LLMs in IP and to support their development as outstanding IP agents for high-quality innovation, we introduce IPEval, the first bilingual benchmark for IP agency and consulting tasks. We collected English and Chinese multiple-choice questions from past patent bar exams administered by the USPTO and China National Intellectual Property Administration (CNIPA). The English data primarily assess knowledge of US patent law, while the Chinese data focus on Chinese patent law and related regulations. This dataset aims to evaluate current LLMs on their understanding of IP-related laws, cognitive recognition of IP activities such as infringement handling and protection methods, and application procedures, as well as their reasoning abilities.

We thoroughly annotated and described this dataset and evaluated the performance of LLMs on IP consulting tasks from multiple perspectives, classifying their performance into different levels. We constructed an evaluation system for IP consulting tasks using four dimensions, eight fields, and five levels. IP consulting tasks require models to have relevant legal knowledge, which is region-specific and evolves over time. Therefore, the evaluation benchmark must incorporate both regional and temporal characteristics\cite{boateng2013hand}.

We used IPEval to evaluate 15 LLMs across 7 types, all of which are bilingual models in Chinese and English. The models were categorized into two groups: 9 open-source models and 6 closed-source models. We applied three prompting strategies: zero-shot\cite{wei2021finetuned}, 5-few-shot\cite{brown2020language}, and 5-few-shot-CoT\cite{wei2022chain}. The results show that most models fell short of the passing mark, except for Qwen-Max\cite{bai2023qwen} and Qwen-72B-Chat\cite{bai2023qwen}, which barely passed with scores of 63.3 and 62.6, respectively, reaching the 2A level.

Overall, there is significant room for improvement in LLMs' capabilities in the IP domain. We hope that IPEval can provide accurate assessments of LLMs' IP capabilities and encourage researchers interested in the IP domain to develop LLMs with richer IP knowledge and stronger reasoning abilities. The contributions of this paper to the community are summarized as follows:
\begin{itemize}
	\item We introduce the first bilingual benchmark IPEval for IP agency and consulting tasks.
 \item Evaluation of the IP capabilities of 15 LLMs using the IPEval benchmark, assessing them across four dimensions, eight fields, and five levels.
 \item To encourage the community to develop LLMs with stronger IP capabilities, we have made the evaluation data and code open-source on github. URL: \url{https://github.com/Mathsion2/IPEval}
\end{itemize}

\section{The IPEval Benchmark}
\subsection{Evaluation System}
We designed an evaluation system for IP agency consulting tasks from multiple perspectives, assessing LLMs across four capability dimensions and eight associated fields. We also considered the benchmark's regional and temporal aspects\cite{boateng2013hand} and categorized the models' capabilities into five levels.

\paragraph{Four capability dimensions:}(1) IP Creation represented by questions on patent application procedures. (2) IP Application represented by questions on methods of patent outcome transformation. (3) IP Protection represented by questions on patent infringement determination. (4) IP Management represented by questions on patent examination analysis. Specific examples and detailed definitions for each dimension are provided in Appendix~\ref{A.1}.

\begin{figure}
	\centering
	\includegraphics[width = \linewidth]{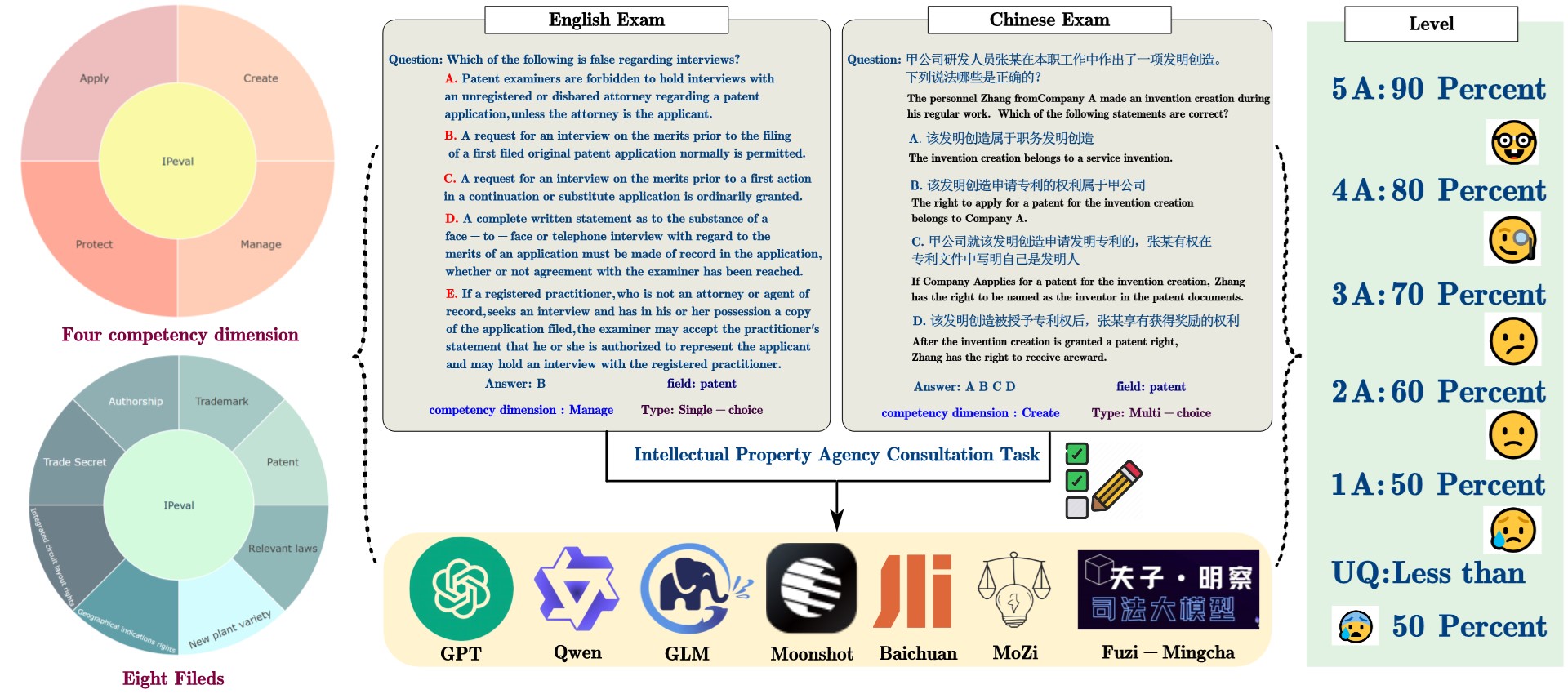}
	\caption{Evaluate system. We have proposed an evaluation system for intellectual property agency consulting tasks~\ref{A.5}, which assesses the model from four capability dimensions and eight knowledge fileds, categorizing the model's performance into five levels.}
	\label{fig:3}
 \end{figure}

\paragraph{Eight fields:}Referring to the World Intellectual Property Organization (WIPO)\cite{davies2020foundation} definition of intellectual property, we constructed data for eight domains, namely: patent rights which including inventions, utility models, and designs, trademark rights, copyrights, trade secrets, integrated circuit layout design rights, geographical indications, and related laws.

\paragraph{Five levels:}We categorize the capabilities of large models in IP agency consulting tasks into five levels: A, 2A, 3A, 4A, and 5A, corresponding to scores of 50, 60, 70, 80, and 90, respectively.

\paragraph{Regional and temporal characteristics:}We collected data from both the USPTO and CNIPA patent bar exams to ensure the benchmark's regional specificity. Additionally, as these exams are conducted annually and updated with regulatory changes, gathering exam questions from multiple consecutive years naturally ensures the benchmark's temporal relevance.

\begin{figure}
  \centering
\includegraphics[width=\linewidth]{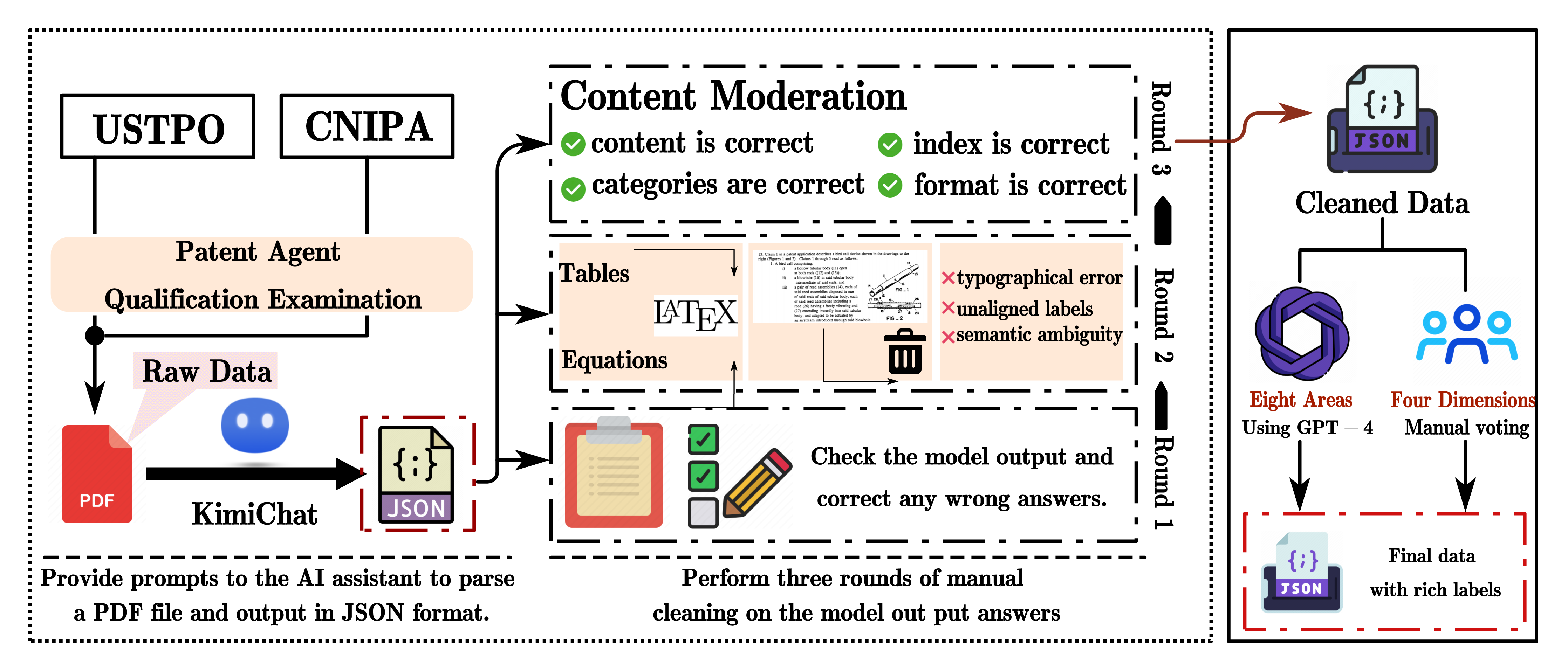}
  \caption{Data source and processing. 
  	We gathered data from the patent agent qualification examinations and carefully cleaned and extensively annotated it. IPEval contains over two thousand high-quality data entries.}
  \label{figure1}
\end{figure}
\subsection{Data Collection}
\label{2.2}
\paragraph{Source:}We collected two sets of data in both Chinese and English\footnote{Our raw data comes from \url{https://wenku.baidu.com/} and \url{https://mypatentbar.com/old-exams/}}: 1) The annual Chinese patent bar exams from 2012 to 2019 and 2) The biannual US patent bar exams from 1997 to 2003. These authentic exam data are publicly available from the USPTO and CNIPA, ensuring the authority, correctness, reliability, and quality of the data. This guarantees the regional and temporal relevance of the dataset and increases the difficulty for models in handling different regional and language issues simultaneously. This ensures a more accurate reflection of the capabilities of LLMs in IP agency and consulting tasks. We assume that LLMs can only make correct judgments when they fully understand different regional regulations.See more details in the Appendix~\ref{A.2}.

\paragraph{Processing:}The source data are all in PDF format, which reduces the risk of data leakage compared to structured text. To efficiently extract the document content, we used KimiChat\footnote{\url{
https://kimi.moonshot.cn/}} to parse PDF documents and requested it to output structured multiple-choice question data in JSON format. To avoid errors from KimiChat, we conducted three rounds of manual verification on the obtained JSON format data, as shown in Figure~\ref{figure1}.
\begin{itemize}
	\item First round: Checked and modified the answer choices for each question outputted by KimiChat against the real answers.
   \item Second round: Further processed reading comprehension and chart analysis questions to convert them into plain text data.
   \begin{itemize}
	\item [1)]Inserted context into the question stem for each reading comprehension question.
   \item [2)]Removed chart-based questions requiring multi-modal capabilities from the model.
   \item [3)]Manually parsed formulas and tables into LaTeX format as replacements.
  \end{itemize}
  \item Third round: Standardized the error extraction symbols in the JSON file and conducted strict content verification.
\end{itemize}

\paragraph{Data annotation:}We conducted detailed annotations on the collected data, including region, time, capability dimensions, and associated fields. Annotations for time and region were easily completed during the data collection stage. We used the following methods for annotating capability dimensions and associated fields:
\begin{itemize}
	\item Capability dimensions: We employed manual annotation methods. A text annotation system was built based on doccano\cite{doccano}, and crowd workers from relevant fields were employed to annotate the data. Statistical results were obtained through voting based on the annotated information.
 \item Associated fields: We utilized GPT-4\cite{achiam2023gpt} for automatic task domain classification, followed by a round of manual verification after GPT-4\cite{achiam2023gpt} annotations were completed.
\end{itemize}
\subsection{Data Statistic}
The collection, cleaning, and design of IPEval required a considerable amount of humanpower. We obtained 2657 multiple-choice questions in both Chinese and English, covering various capability dimensions and associated fields. The dataset aims to evaluate the capabilities of large models in understanding, applying, and reasoning tasks related to IP. We provided a detailed description of the dataset from multiple perspectives.

The assessment format is multiple-choice questions, which requires the model to have a more accurate understanding of knowledge. Compared with subjective questions that rely on text generation, it can eliminate the subjective influence of evaluators. In IPEval, the data distribution of single-choice and multiple-choice questions, options, and ability dimensions is balanced, which helps to reduce potential biases. The Chinese data contains four options and two types, the English data has five options.

\begin{figure}
  \centering
\includegraphics[width=\linewidth]{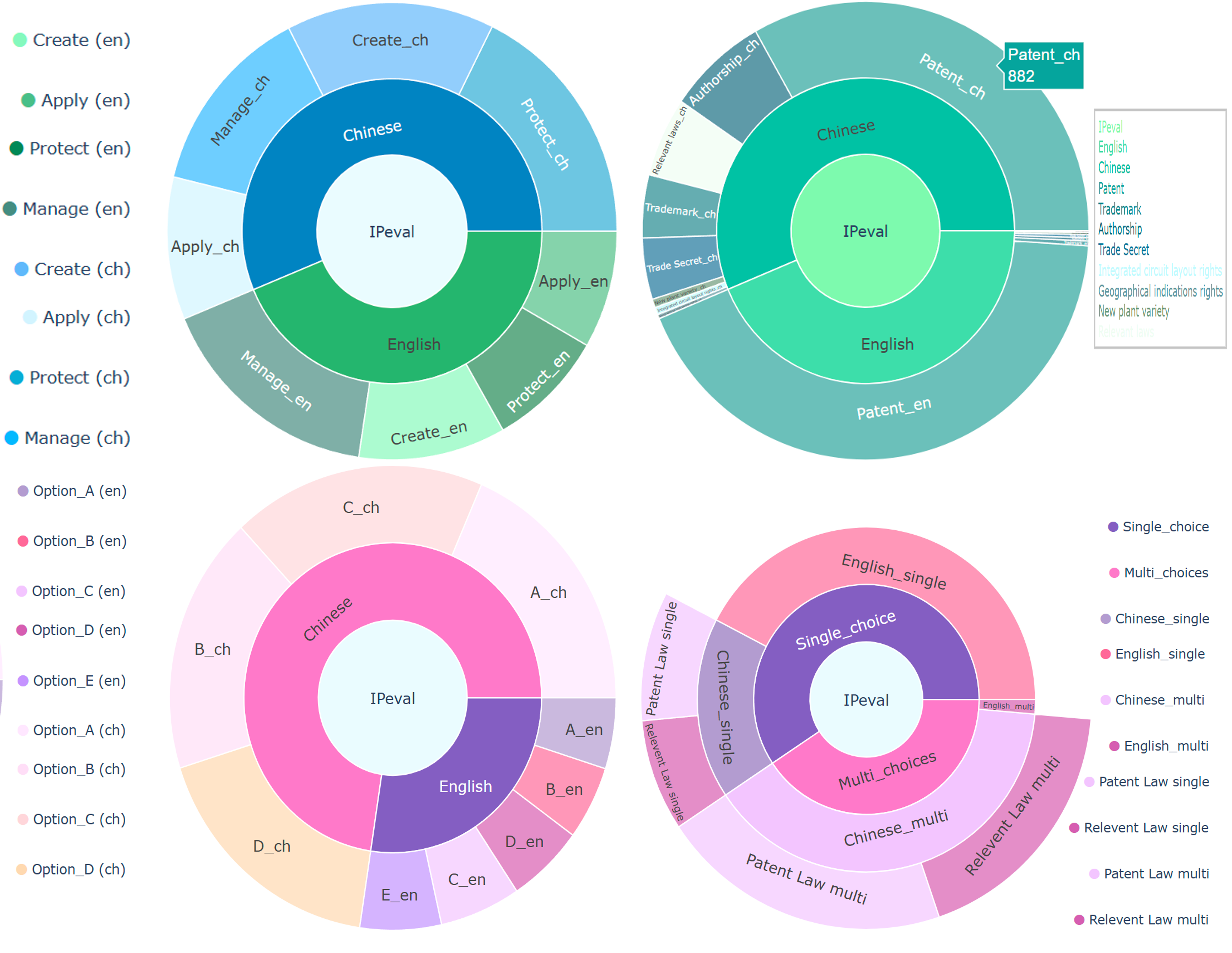}
  \caption{Data statistic. We conducted multi-dimensional statistical analysis and categorization of the data, including language (Chinese and English), four capability dimensions, eight knowledge Fields, distribution of correct answers, distribution of single and multiple-choice questions, as well as the distribution of Chinese data related to patent law and relevant regulations. For more information, please refer to the Appendix~\ref{A.2}.}
  \label{figure2}
\end{figure}

Patents are the main carriers of IP, reaching 97.59\% in English data .We refer to WIPO's classification standards\cite{davies2020foundation} for intellectual property rights to divide the fields, as shown in Figure ~\ref{figure2}.

\section{Experiment}
\label{3}
\subsection{Prompt}
We set up similar but not entirely identical prompts for different languages, question types, and experimental types. For Chinese, we introduced the following sentence before each question: "请你作为一个专利代理律师，以下是中国专利代理人资格考试的\{choice\_type\}。请以下列形式回答，答案：". In this case, to alleviate the model's burden, prompts for single-choice and multiple-choice questions were differentiated based on the number of answer choices. For single-choice questions, \{choice\_type\} was specified as: "单项选择题，请选出四个选项中最符合题目要求的一个答案". For multiple-choice questions, \{choice\_type\} was specified as: "多项选择题，请选出四个选项中所有符合题目要求的答案"。

For English, we introduced the following sentence before each question: "You are a patent attorney, the following is a multiple-choice question from the USPTO Patent Attorney Exam, \{choice\_type\}." Similar to the Chinese prompts, prompts for single-choice and multiple-choice questions were differentiated based on the number of answer choices. For single-choice questions, the prompt was: "Select only one the best answer for this question. The format of the answer should be 'Answer: option.'" For multiple-choice questions, the prompt was: "Select all the answers for the question. The format of the answer should be 'Answer: option1, option2, ...'".

For few-shot\cite{brown2020language} experiments, we chose to provide 5 examples for each test question as additional supplementary information, known as 5-few-shot. We ensured that the test questions did not appear in the examples to avoid answer leakage. Building upon 5-few-shot, for Chinese questions, we introduced "让我们一步一步思考。" placed after the last example and before the test examples. For English questions, in addition to the "Let's think step by step" prompt, we added extra answer explanations for each example, forming a triplet of question, analysis (thought process) and answer. The full form of prompts for zero-shot, few-shot, and CoT can be found in Appendix~\ref{A.3}.

\begin{table}[htbp]
	\centering
	\caption{The models evaluated in this paper. The models marked with an asterisk (*) are LLMs in vertical domains.}
	 \scalebox{0.8}{\begin{tblr}{
			cells = {c},
			hline{1,17} = {-}{0.08em},
			hline{2} = {-}{},
		}
		\textbf{Model}                     & \textbf{Creator} & \textbf{Parameters} & \textbf{Access} \\
		ChatGPT\cite{achiam2023gpt}             & OpenAI           & undisclosed                  & API             \\
		GPT-4\cite{achiam2023gpt}                              & OpenAI           & undisclosed                  & API             \\
		moonshot-v1-8k                     & Moonshot         & undisclosed                  & API             \\
		Qwen-7B-Chat\cite{bai2023qwen}                       & Alibaba          & 7B                  & API             \\
		Qwen-72B-Chat\cite{bai2023qwen}                      & Alibaba          & 72B                 & API             \\
		Qwen1.5-14B-Chat\cite{bai2023qwen}                   & Alibaba          & 14B                 & API             \\
		Qwen-Max\cite{bai2023qwen}                           & Alibaba          & undisclosed                  & API             \\
		Baichuan2-7B-Chat\cite{yang2023baichuan}                  & Baichuan         & 7B                  & API             \\
		Baichuan2-13B-Chat\cite{yang2023baichuan}                 & Baichuan         & 13B                 & API             \\
		ChatGLM3-6B\cite{du-etal-2022-glm}                        & ZhiPu         & 6B                  & Weights         \\
		GLM-3-turbo\cite{zeng2022glm}                        & ZhiPu         & 130B                & API             \\
		GLM-4\cite{zeng2022glm}                               & ZhiPu         & 130B                & API             \\
		fuzi-mingcha-v1\textsuperscript{*}\cite{deng-etal-2023-syllogistic} & SDUIRLab         & 6B                  & Weights      \\ 
        MoZi-glm\textsuperscript{*}\cite{ni2024mozip} & SIAT & 6B & Weights \\
        MoZi-bloomz\textsuperscript{*}\cite{ni2024mozip} & SIAT & 7B & Weights \\
	\end{tblr}}
	\label{Table1}
\end{table}
\subsection{Models}
As shown in Table~\ref{Table1}, we selected 7 kinds of different Chinese and English LLMs with varying parameters and generalizability for testing. In the general domain, we chose closed-source models such as ChatGPT\cite{achiam2023gpt}, GPT-4\cite{achiam2023gpt}, Qwen-max\cite{bai2023qwen}, GLM-4\cite{zeng2022glm}, and moonshot-v1-8k, most of which have parameter counts exceeding hundreds of billions; and open-source models like ChatGLM3-6B\cite{du-etal-2022-glm}, Qwen1.5-14B-Chat\cite{bai2023qwen}, with less parameter. In vertical domains, we evaluated the legal LLM Fuzi-Mingcha\cite{deng-etal-2023-syllogistic} and the IP LLM MoZi\cite{ni2024mozip}. Fuzi-Mingcha, derived from the ChatGLM-6B base model fine-tuned, claimed to outperform models like ChatLaw\cite{cui2023chatlaw} and LawGPT\cite{LAWGPT-zh} in the legal domain. Mozi comes in two variants: ChatGLM3-6B and Bloomz-MT-7B\cite{muennighoff2023crosslingual}, capable of handling IP dialogue QA and patent matching tasks in nine languages. For more detailed information on these models, see Appendix~\ref{B}.
\subsection{Main Results}
As shown in Table~\ref{Table2}, only Qwen-Max and Qwen-72B-Chat reached the 2A level, while GPT-4, Qwen1.5-14B-Chat, and GLM-4 reached the A level. No model achieved the 3A level in the test. Overall, Qwen-Max performed the best, demonstrating good abilities in IP consulting tasks. Models with larger parameters generally performed better in the test, showing stronger reasoning and IP consulting abilities. However, in our evaluation system, none of the models reached satisfactory levels of 4A or 5A. The overall performance of large models was generally poor, even those with higher parameter counts only reached the 2A level. We provided a more detailed description and evaluation of the model's performance from various perspectives such as different domains, ability dimensions, and source, which shown in Appendix~\ref{C}. 

\begin{table}[htbp]
	\centering
	\caption{Main results. The table contains the test results of different types of large models on all data, using three prompt strategies. "UQ" represents unqualified, failing to reach the minimum requirement of grade A.}
	\scalebox{0.8}{\begin{tblr}{
			cells = {c},
			hline{1,17} = {-}{0.08em},
			hline{2,4,5,9,11,14} = {-}{},
		}
		\textbf{Model}                     & \textbf{Zero-Shot} & \textbf{5-Few-Shot} & \textbf{5-Few-Shot+CoT} & \textbf{Average} &\textbf{Level}\\
		gpt-3.5-turbo(ChatGPT)             & 33.1               & 33.5                & 32.7                    & 33.1     &UQ        \\
		GPT-4                              & 55.7               & 55.5                & 54.7                    & 55.3      &A       \\
moonshot-v1-8k& 45.5               & 46.6                & 46.1                    & 46.1      &UQ       \\
		Qwen-7B-Chat                       & 33.9               & 29.8                & 27.4                    & 30.3       &UQ      \\
		Qwen-72B-Chat                      & 61.5               & \textbf{64.5}       & 61.9                    & 62.6        &2A     \\
		Qwen1.5-14B-Chat                   & 52.1               & 52.1                & 49.4                    & 51.2     &A        \\
		Qwen-Max                           & \textbf{62.4}      & 64.2                & \textbf{63.4}           & \textbf{63.3}  &\textbf{2A}  \\
		Baichuan2-7B-Chat                  & 23.9               & 28.2                & 24.7                    & 25.6      &UQ       \\
		Baichuan2-13B-Chat                 & 32.9               & 11.9                & 19.3                    & 21.4       &UQ      \\
		ChatGLM3-6B                        & 23.5               & 21.6                & 26.7                    & 23.9       &UQ      \\
		GLM-3-turbo                        & 50.2               & 50.2                & 48.2                    & 49.5       &UQ      \\
		GLM-4                              & 55.9               & 58.1                & 57.8                    & 57.3     &A        \\
		fuzi-mingcha-v1\textsuperscript{*} & 17.1               & 15.3                & 9.2                     & 13.9             &UQ
\\
  MoZi-glm\textsuperscript{*}  &9.98  & 10.69 & 3.29 &7.99  &UQ
\\
  MoZi-bloom\textsuperscript{*} &8.30  & 3.67 &1.90  &4.62  &UQ\\
\end{tblr}}
	\label{Table2}
\end{table}

\section{Analysis}
\subsection{Performance Analysis}

Our analysis of model performance from different linguistic perspectives, as detailed in Appendix~\ref{D}, reveals that models primarily using English or Chinese excel in corresponding tasks, demonstrating a regional preference in IP consulting tasks. We also analyzed performance based on the number of answer options, as shown in Appendix~\ref{D.1}. The results indicate that in Chinese tests, all models perform significantly better on single-choice questions than on multiple-choice questions, with this effect being particularly pronounced in smaller models. For example, Baichuan2-7B-Chat, fuzi-mingcha-v1*, and MoZi* achieve less than 10\% accuracy in multiple-choice questions. This discrepancy highlights the greater difficulty of multiple-choice questions, which require LLMs to possess enhanced reasoning capabilities.

From the perspective of evaluation methods, we assessed the impact of few-shot and CoT approaches on model performance, as shown in Appendix~\ref{D.1}. Most models significantly improved under the 5-few-shot strategy; for instance, Baichuan2-7B-Chat saw a 38.6\% performance increase in Chinese questions with this approach, while Qwen-72B-Chat improved by 9.3\%, achieving a high score of 77.3, surpassing the larger Qwen-Max model. However, some models experienced a drastic decline in performance after adopting this strategy, such as Baichuan2-13B-Chat, which saw a 63.8\% drop, performing even below its smaller counterpart, Baichuan2-7B-Chat.

Almost all models showed a performance decrease when CoT was added on top of the 5-few-shot basis, except for ChatGLM3-6B, which improved by 23.6\%. Baichuan2-13B-Chat, after incorporating CoT, dropped by 41.3\% compared to zero-shot but gained 62.2\% over its few-shot performance (its zero-shot performance was superior to few-shot). In contrast to Baichuan2-13B-Chat, larger models such as GLM-4, Qwen-Max, and Moonshot-v1-8K all showed performance improvements with the introduction of CoT compared to zero-shot. The performence of CoT and other patterns are shown in Figure \ref{zf}, \ref{zc}, \ref{fc}.

\subsection{Error Analysis}
\label{4.2}
We meticulously examined the specific outputs generated by models in our experiments, identifying various types of errors. One common mistake involved inconsistent reasoning: in the preliminary thought process, the model erroneously dismissed option C and chose it, yet later concluded that option D was the correct answer. Another error type occurred when models failed to understand the conditions stipulated in the question stem, leading to outputs that perfectly avoided the correct answers, particularly in multiple-choice tasks.

Additionally, errors arose from the refusal to answer questions due to ethical or political censorship. LLMs trained with reinforcement learning from human feedback (RLHF)\cite{bai2022training} are adept at rigorously assessing whether user inputs violate ethical standards or involve political content. In our study, given that the introduced benchmarks include legal content, certain keywords triggered content filters, leading to refusal of responses. For example, the Qwen model provided by the Dashscope platform on Alibaba Cloud returns a 400 error code when it detects inappropriate content, stating "Input or output data may contain inappropriate content." It is important to clarify that the IPEval benchmark does not contain any content that breaches ethical standards or involves sensitive political issues, as shown in Appendix~\ref{D.2}, Figure~\ref{ethnic}.

Apart from model capabilities, errors due to overly long contexts in 5-few-shot or CoT scenarios were observed. Models like fuzi·mingcha and MoZi were unable to perform normally in CoT experiments, as their context window is limited to 2048 tokens, and the average token length in English CoT experiments exceeded this limit, preventing the models from fully comprehending the text.

\section{Conclusion}

LLMs have vast potential for development in the field of IP, aiding in addressing the scarcity of consulting services. Evaluating the performance of LLMs in IP consulting tasks contributes to their further development. We introduces the IPEval benchmark, collecting patent attorney exam questions from the USTPO and CNIPA, and constructs an assessment system based on multiple domains and ability dimensions. By using multiple-choice questions, we can accurately assess the model's ability in IP consulting tasks, eliminating subjective biases and better reflecting the model's precise understanding and application of relevant knowledge. We also conducted a detailed analysis of the model's performance in terms of locality and timeliness. We do not intend for IPEval to be a competitive ranking, but rather to lay the foundation for the vertical development of large language models in the field of IP research and development. The contributions we made are as following:

\begin{itemize}
	\item We provide a rigorous definition of IP consulting tasks and propose the first evaluation system tailored to IP consulting tasks.。
   \item We discuss in detail the timeliness and locality of IP datasets, as well as how they ensure geographical and temporal relevance.
   \item We find models primarily using Chinese, such as the Qwen series, outperform powerful general-purpose models like GPT-4 in the Chinese IP domain tasks,.
   \item We find some specially trained and fine-tuned legal domain LLMs still lag behind general-purpose LLMs in IP performance, indicating the necessity and significant potential for developing more specialized IP domain LLMs. 
\end{itemize}
 Overall, compared to Zero-Shot, models perform slightly better in Few-Shot and CoT, but CoT's performance is slightly worse than Few-Shot. The experimental settings of Few-Shot and CoT have a stronger positive guiding effect on multiple-choice questions. We have open-sourced the IPEval dataset on github to facilitate researchers in using it as training data or evaluation benchmarks for further model optimization and evaluation.

\begin{ack}
This work is partially supported by grant from the Natural Science Foundation of China (No. 61976036).
\end{ack}



\appendix
\section{The Detail of IPEval}

\subsection{Data Examples}
\label{A.1}
\begin{longtable}[htbp]{>{\centering\arraybackslash}m{0.15\linewidth} m{0.85\linewidth}} 
	
	\caption{Examples of data samples for the four capability dimensions. The dimensions include: Creation, Application, Protection, and Management.}\\ 
	
	\toprule 

	\textbf{Question} & \textcolor[rgb]{0.051,0.051,0.051}{Company A completed the creation of a certain layout design on June 10, 2018, first put it into commercial use on February 1, 2019, applied for registration with the National Intellectual Property Administration on August 12, 2019, and was granted registration on September 12, 2019. According to the regulations on the protection of integrated circuit layout designs, from which date should the protection period for this layout design begin?}\\ 
	
	\midrule 
	
	\textbf{Options} & A. 2018-6-10~ \textbf{B. 2019-2-1}~ C. 2019-8-12 ~D. 2019-9-12\par\textbf{Label
		is: Creation} \\ 
	\hline\hline 
	
	\textbf{Question} & \textcolor[rgb]{0.051,0.051,0.051}{Screenwriters Wang and Li jointly created an indivisible movie script. Regarding whether to permit or transfer the filming rights of the script, Wang and Li have differing opinions. Li disagrees with both licensing and transferring without just cause. According to copyright law and related regulations, which of the following statements is correct?}  \\ 
	
	\midrule 
	
	\textbf{Options} & A. Wang cannot permit others to film the script into a movie.\par \textbf{B. Wang can permit others to film the script into a movie, and the proceeds belong to Wang alone.}\par C. Wang can permit others to film the script into a movie, but the proceeds should be reasonably distributed with Li.\par D. Wang can transfer the filming rights of the script to others, but the proceeds should be reasonably distributed with Li. \par\textbf{Label is: Application} \\ 
	\hline\hline 
	
	\textbf{Question} & \textcolor[rgb]{0.051,0.051,0.051}{Independently of each other, M and Z invented the same tennis racket in the United States. A U.S. patent was granted to M on February 18, 1997, on an application filed on April 12, 1995 claiming the same tennis racket. On March 10, 1997, Z filed a patent application in the PTO claiming the same tennis racket. There is no common assignee between M and Z, and they are not obligated to assign their inventions to a common assignee. M and Z filed their respective patent applications diligently shortly after each conceived the tennis racket invention. Under which of the following provisions of 35 U.S.C § 102 is the U.S. patent to M prior art with regard to the tennis racket claimed by Z?} \\ 
	
	\midrule 
	
	\textbf{Options} & A. 35 U.S.C § 102(a)~~ B. 35 U.S.C § 102(e) ~~C. 35 U.S.C § 102(g) \par D. (A) and (B) only. ~~\textbf{E. (A), (B), and (C).}\par\textbf{Label is: Protection} \\ 
	\hline\hline 
	
	\textbf{Question} & \textcolor[rgb]{0.051,0.051,0.051}{On April 4, 1997, practitioner P filed a patent application for inventor S claiming a laminate of several layers. The examiner rejected the claims in the application as being anticipated under 35 U.S.C. § 102(a) and 102(e) over a U.S. patent granted on February 11, 1997, to Jones which describes the laminate in the specification, and how to make the laminate. Jones did not claim the laminate, or derive the laminate from S. The patent was granted to Jones on an application filed on November 2, 1994. The claims in the Jones patent are directed to a substrate having a single layer. S first conceived his invention on July 4, 1995. Which of the following would be (an) appropriate response(s) to overcome the rejection?} \\ 
	
	\midrule 
	
	\textbf{Options} & \textbf{A. Amending the claim by adding limitations which are supported by the specification in S's application, and are not disclosed or suggested by the Jones patent, request reconsideration, and clearly point out the patentable novelty of the amended claim presented over Jones.}\par B. Produce evidence of secondary considerations such as unexpected results or commercial success.\par C. Argue that the claimed invention in the Jones patent is non-analogous art or teaches away from the invention.\par D. Argue that the Jones patent is not evidence that the invention was known "by others," as required by 35 U.S.C. § 102(a) since Jones is a single person, whereas "others" requires plural persons.\par E. B and C. \par\textbf{Label is: Management} \\ 
	
	\bottomrule 
\end{longtable} 

\begin{longtable}[htbp]{>{\centering\arraybackslash}m{0.15\linewidth} m{0.85\linewidth}} 
	
	\caption{Examples of data samples for the eight knowledge domains. Including: Patent, Trademark, New plant variety, Trade Secret, Integrated circuit layout rights, Relevant laws, Authorship, Geographical indications rights. }\\ 
	
	\toprule 

	\textbf{Question} & On March 20, 1997, inventor B filed a patent application in the PTO claiming invention X. Inventor B did not claim priority based on a foreign application filed by inventor B on June 3, 1996, in the Patent Office of Germany. In the foreign application, inventor B disclosed and claimed invention X, which inventor B had conceived on October 10, 1995, and reduced to practice on January 5, 1996, all in Germany. The patent examiner issued an Office action where all the claims in the patent application were properly rejected under 35 U.S.C. § 102(a) and (e) as being anticipated by a U.S. patent granted to inventor Z on November 5, 1996, on a patent application filed in the PTO on February 6, 1995. There is no common assignee between Z and B, and they are not obligated to assign their inventions to a common assignee. Moreover, inventor Z, independently of inventor B, invented invention X, and did not derive anything from inventor B. The U.S. patent discloses, but does not claim, invention X. Which of the following is/are appropriate response(s) which could overcome the rejections under § 102(a) and (e) when timely filed?\\ 
	
	\midrule 
	
	\textbf{Options} & \textbf{A. Amend the claims to require particular limitations disclosed in inventor B's application, but not disclosed or suggested in inventor Z's patent, and argue that the limitations patentably distinguish the claimed invention over the prior art.}\par 
	B. File an antedating affidavit or declaration under 37 CFR § 1.131 showing conception on October 10, 1995, and reduction to practice on January 5, 1996, all in Germany. \par
	C. File a claim for the right and benefit of foreign priority wherein the German application is correctly identified, file a certified copy of the original German patent application, and argue that as a result of the benefit of foreign priority the U.S. patent is no longer available as a prior art reference against the claims. \par
	D. (A), (B) and (C).~~~~~~E. (A) and (C) only.\par
	\textbf{Label is: Patent} \\ 
	\hline\hline 
	
	\textbf{Question} & Which of the following constitutes a proper method of service to an unrepresented applicant of papers in a protest of an application?  \\ 
	
	\midrule 
	
	\textbf{Options} & A. Transmission by first class mail to the applicant.
	\par B. Personally delivering a copy to the applicant.\par 
	C. Leaving a copy at the usual place of business of the applicant with someone in the applicant's employment.\par 
	D. When there is no usual place of business by leaving a copy at the applicant's home with someone of suitable age and discretion who resides there.\par
	\textbf{E. All of the above.}\par 
	\textbf{Label is: Trademark} \\ 
	\hline\hline 
	
	\textbf{Question} & You have filed a complete plant patent application claiming 1) a distinct and new plant variety and 2) a method for obtaining the plant variety. Which of the following statements is/are false?I. You may not amend the application to add additional description of the plant variety inadvertently omitted from the original application. II. You may be required to deposit an adequate sample of the plant variety with an acceptable depository and the claims may be rejected under 35 U.S.C. § 112 without the deposit. III. You may be required to restrict the claims between plant variety and plant method inventions you want examined for ultimate issuance as the single claim in the plant patent application to which you are entitled. \\ 
	
	\midrule 
	
	\textbf{Options} & A. III ~~~B. II and III~~~\textbf{C. I and II}~~~ D. I and III~~~ E. I, II, and III.
	\par\textbf{Label is: New plant variety} \\ 
	\hline\hline 
	
	\textbf{Question} & You file a patent application on behalf your client, a small company, claiming a cleaning solution wherein uric acid is the primary ingredient. You received an Office action dated August 19, 1997, wherein all the claims were rejected under 35 U.S.C. § 112. You provided your client with a copy of the Office action. Today, August 27, 1997, the President of the client called and informed you that the company wants to keep the composition of the uric acid-containing cleaning solution a trade secret. The President informed you that the company has developed another cleaning solution composition wherein sulfuric acid is the primary ingredient, and uric acid is not present. The President also informed you that the sulfuric acid-containing cleaning solution is better able to clean than the uric acid-containing cleaning solution. The use of sulfuric acid was not disclosed in the pending application. Which of the following is in accord with PTO practice and procedure to obtain patent protection for the sulfuric acid-containing cleaning solution and maintain the confidentiality of the uric acid-containing cleaning solution?
	\\ 
	
	\midrule 
	
	\textbf{Options} & \textbf{A. Abandon the pending application, and file a new patent application disclosing and claiming only the sulfuric acid-containing cleaning solution wherein sulfuric acid is the primary ingredient. The new application should not refer back to the previous, pending application.}\par 
	B. File a continuation-in-part application disclosing and claiming the sulfuric acid-containing cleaning solution wherein sulfuric acid is the primary ingredient. The first sentence of the specification of the continuation-in-part application should refer to the previously filed, copending application.\par 
	C.File an amendment in the pending application adding claims to the sulfuric acid-containing cleaning solution wherein sulfuric acid is the primary ingredient.\par
	D. File a reissue application on the pending application and broaden the claims in the reissue application to be inclusive of the sulfuric acid-containing cleaning solution wherein sulfuric acid is the primary ingredient.\par 
	E. File an amendment in the pending application directing that the word 'uric' be replaced with the word 'sulfuric' wherever 'uric' occurs in the specification and claims. \par
	\textbf{Label is: Trade Secret} \\ 
	
	\bottomrule 
	\textbf{Question} & According to the Regulations on the Protection of Layout Designs of Integrated Circuits and relevant provisions, in the case of a dispute arising from the infringement of exclusive rights to a layout design, to which department can the layout design rights holder or interested party request for handling?\\ 
	
	\midrule 
	
	\textbf{Options} & A. State Administration for Industry and Commerce under the State Council.\par 
	\textbf{B. State Intellectual Property Office under the State Council.}\par 
	C. Science and Technology Administration under the State Council.\par
	D. Provincial-level Department in Charge of Patent Administration.\par\textbf{Label
		is: Integrated circuit layout rights} \\ 
	\hline\hline 
	
	\textbf{Question} & \textcolor[rgb]{0.051,0.051,0.051}{After accepting Wang's application for administrative reconsideration, an administrative reconsideration body found that Wang had already filed an administrative lawsuit with the People's Court, which had been accepted before applying for administrative reconsideration. According to the Administrative Reconsideration Law and relevant provisions, how should the administrative reconsideration body handle this situation?}\\ 
	
	\midrule 
	
	\textbf{Options} & A. Notify the People's Court to suspend the trial of the administrative lawsuit.\par 
	B. Transfer the case to the People's Court for joint trial.\par 
	C. Suspend the review and continue after the People's Court has made a final judgment.\par 
	\textbf{D. Dismiss the application for administrative reconsideration.}\par 
	\par\textbf{Label
		is: Relevant laws} \\ 
	\hline\hline 
	
	\textbf{Question} & \textcolor[rgb]{0.051,0.051,0.051}{According to the Copyright Law and relevant provisions, which of the following belongs to the economic rights within copyright?}\\ 
	
	\midrule 
	
	\textbf{Options} &A. The right to decide whether to make the work available to the public.\par 
	B. The right to authorize others to modify the work.\par 
	\textbf{C. The right to publicly display copies of the artwork.}\par 
	D. The right to indicate the author's identity and have the author's name on the work.\par 
	\par\textbf{Label is: Authorship} \\ 
	\hline\hline 
	
	\textbf{Question} & \textcolor[rgb]{0.051,0.051,0.051}{"CiXi Honey Pear" is produced in CiXi City, Ningbo, Zhejiang Province. It is famous for being different from ordinary pears, with larger fruit, exceptionally abundant water content, and extremely rich nutrition. Now, the region wants to register and protect the geographical indication of "CiXi Honey Pear." How can they apply for registration under the trademark law system?}\\ 
	
	\midrule 
	
	\textbf{Options} & A. Apply for registration of a regular trademark.\par 
	B. Apply for registration of a geographical indication trademark.\par 
	C. Apply for registration of a combined trademark.\par 
	\textbf{D. Apply for registration of a certification trademark.}\par\textbf{Label is: Geographical indications rights} \\ 
	\hline\hline 

\end{longtable} 

\subsection{Data Statistic}
\label{A.2}
For more details, please see table\ref{tab:table1}, \ref{tab:table2}, \ref{tab:table3}, \ref{tab:table4}, \ref{tab:table5}.We presented the data from multiple perspectives.
\begin{table}[htbp]
	\centering
	\caption{The distribution of single-choice and multiple-choice questions in IPEval. We conducted statistical analysis from both Chinese and English perspectives.}
	\begin{tabular}{ccc} 
		\toprule
		& \textbf{Single-Choice} & \textbf{Multiple-Choice}  \\ 
		\midrule
		\textbf{Total} & 59.4\%(1577)           & 40.6\%(1080)              \\
		Chinese & 30.2\%(451)        & 69.8\%(1044)   \\
		English & 96.9\%(1126)       & 3.1\%(36)      \\
		\bottomrule
	\end{tabular}
	\label{tab:table1}
\end{table}

\begin{table}[htbp]
	\centering
	\caption{The quantity statistics of the four capability dimensions data on IPEval. We conducted statistical analysis from both Chinese and English perspectives.}
	\begin{tabular}{cccll} 
		\toprule
		& \textbf{Creation} & \textbf{Application} & \textbf{Protection} & \textbf{Management}  \\ 
		\hline
		\textbf{Total} &    677(25.5\%)         &     488(18.4\%)        &   696(26.2\%)          &     796(29.9\%)         \\
		Chinese	&	397(26.6\%)	&268(17.9\%)	&469(31.4\%)	&361(24.1\%)	\\
		English	&	281(24.2\%)	&	221(19.0\%)	&	225(19.4\%)	&	435(37.4\%) \\
		\bottomrule
	\end{tabular}
	\label{tab:table2}
\end{table}

\begin{table}[htbp]
	\centering
	\caption{The quantity statistics of Chinese and English data, as well as the statistics of two types of data on patent law and related laws in Chinese.}
	\begin{tblr}{
			cells = {c},
			cell{1}{1} = {c=2}{},
			cell{2}{3} = {r=2}{},
			hline{1,4} = {-}{0.08em},
			hline{2} = {-}{},
			hline{3} = {1-2}{},
		}
		\textbf{Chinese } &             & \textbf{English} \\
		Patent law               & Relevant law          & 1162(43.7\%)     \\
		797(30.0\%)       & 698(26.3\%) &                  
	\end{tblr}
	\label{tab:table3}
\end{table}

\begin{table}[htbp]
	\centering
	\caption{The distribution of correct answers.}
	\begin{tblr}{
			cells = {c},
			hline{1,7} = {-}{0.08em},
			hline{2} = {-}{},
		}
		\textbf{Option} & \textbf{Chinese} & \textbf{English} \\
		A               & 25.5\%(855)           & 18.1\%(228)           \\
		B               & 24.8\%(831)           & 18.7\%(237)           \\
		C               & 25.4\%(851)           & 21.4\%(271)           \\
		D               & 24.4\%(820)           & 20.3\%(257)           \\
		E               & -                & 21.5\%(272)           
	\end{tblr}
	\label{tab:table4}
\end{table}

\begin{table}
	\centering
	\caption{Eight fileds detail. PA: Patent, TM: Trademark,AU:  Authorship, TS: Trade Secret, IC: Integrated circuit layout rights, GE: Geographical indications rights, NP: New plant variety, RL: Relevant laws. \\}
	\resizebox{13.5cm}{!}{
		\begin{tblr}{
				cells = {c},
				hline{1,5} = {-}{0.08em},
				hline{2} = {-}{0.05em},
			}
			& PA        &TM & AU  & Trade Secret & IC & GE & NP & RL \\
			Chinese & 882(59.0\%)   & 118(7.9\%) & 193(12.9\%) & 115(7.7\%)   & 14(0.9\%)                        & 8(0.5\%)                        & 16(1.1\%)         & 149(10.0\%)   \\
			English & 1134(97.59\%) & 9(0.77\%)  & 6(0.52\%)   & 6(0.52\%)    & 2(0.17\%)                        & 1(0.09\%)                       & 4(0.34\%)         & 0(0\%)        \\
			Total   & 2016(75.9\%)  & 127(4.8\%) & 199(7.5\%)  & 121(4.6\%)   & 16(0.6\%)                        & 9(0.3\%)                        & 20(0.7\%)         & 149(5.6\%)    
		\end{tblr}
	}
	\label{tab:table5}
\end{table}

\begin{figure}[htbp]
	\centering
	\includegraphics[width = \linewidth]{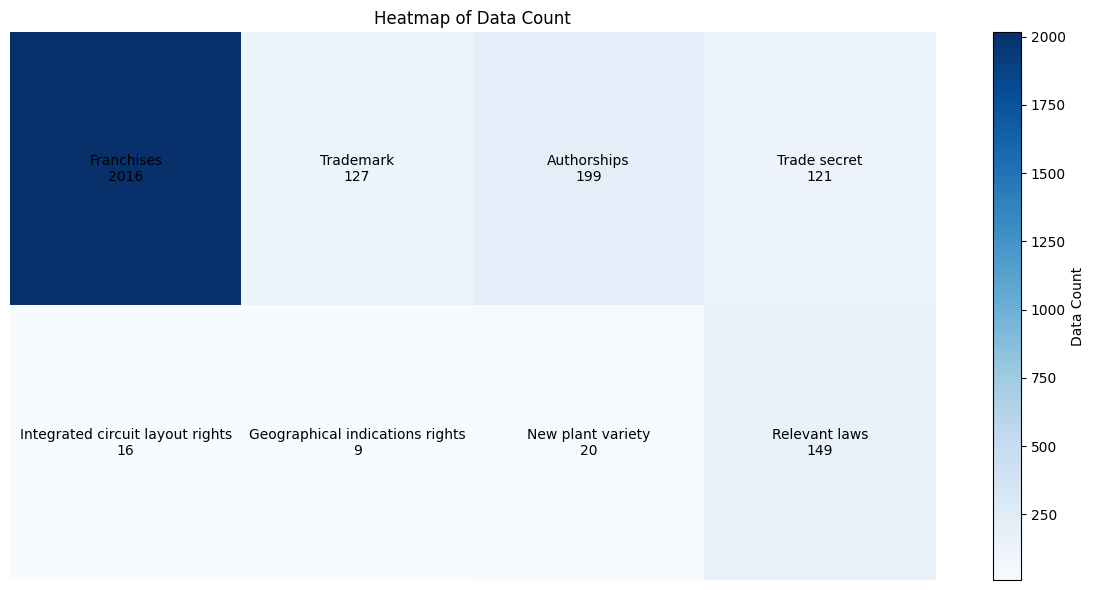}
	\caption{A heat map of the quantity of data for the eight knowledge fileds.}
	\label{heatmap}
\end{figure}

\subsection{Complete Prompt}
\label{A.3}
\begin{figure}[htbp]
	\centering
	\includegraphics[width = \linewidth]{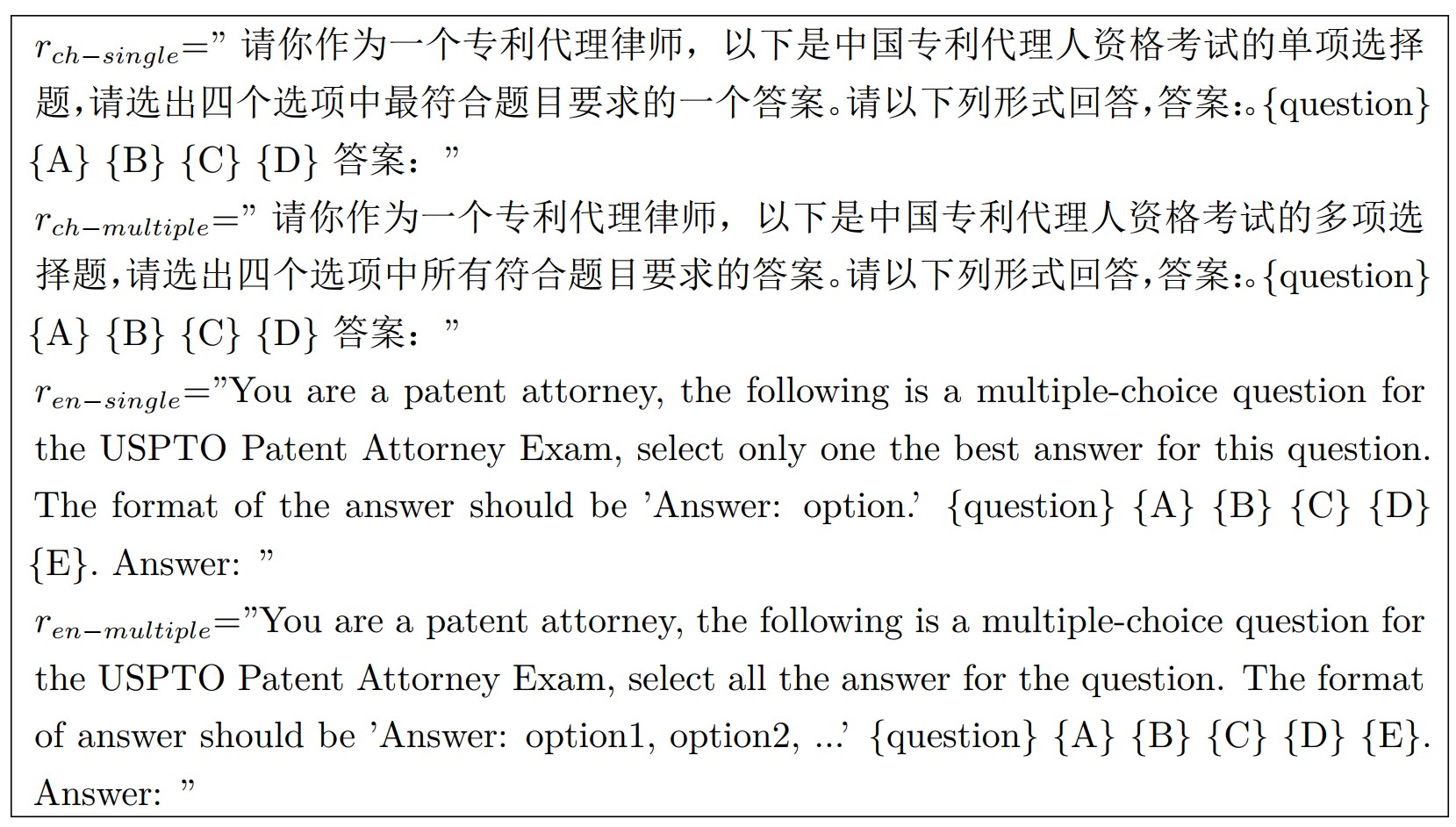}
	\caption{Complete prompt. For Chinese questions, use Chinese prompts; for English questions, use English prompts.}
	\label{prompt}
\end{figure}

The complete prompt can be seen in figure \ref{prompt}.

\subsection{Average Length of Data}
\label{A.4}
The table \ref{tab:table11} presents the average context lengths for different language tasks and experiment types. It demonstrates that the average context length for Chain of Thought experiments in both Chinese and English is approximately five times longer than that of the Zero-Shot experiments. Moreover, the average context length for Chain of Thought experiments in English exceeds 3,000 tokens, surpassing the maximum acceptable context length of 2048 tokens for the fuzi-mingcha-v1$^*$ model in this experimental setting. 

\begin{table}[htbp]
	\centering
	\caption{Average length of prompt/token.}
	\scalebox{0.7}{
		\begin{tblr}{
				row{1} = {c},
				row{3} = {c},
				cell{1}{2} = {c=3}{},
				cell{1}{5} = {c=3}{},
				cell{2}{1} = {c},
				cell{2}{2} = {c},
				cell{2}{5} = {c},
				hlines,
				vline{3} = {1}{},
				vline{5} = {2-3}{},
				hline{1,4} = {-}{0.08em},
			}
			\textbf{Language}             & \textbf{Chinese } &            &                & \textbf{English } &            &                \\
			\textbf{Experiment Pattern}   & Zero-Shot         & Few-Shot & Few-Shot+CoT & Zero-Shot         & Few-Shot & Few-Shot+CoT \\
			\textbf{Average Length/token} & 131.05            & 648.09     & 649.72         & 306.959           & 1669.127   & 2935.79             
		\end{tblr}
	}
	\label{tab:table11}
\end{table}

\subsection{Task Formulation}
\label{A.5}
Intellectual Property Agency Consulting Task: Refers to the rights holder making inquiries during the validity period of intellectual property rights (from the beginning of the intellectual property application) regarding the application process, review period, rights transfer, outcome conversion, legal consultation, and infringement judgment series of matters concerning the large-scale model.

The scope of intellectual property consulting tasks is extremely broad, covering multiple fields such as law and technology, making the assessment of intellectual property tasks very difficult. To address this, we have designed an evaluation system for intellectual property consulting tasks from multiple perspectives. This system evaluates the large-scale model based on four dimensions of capability and eight associated fields, while also considering the benchmark's regional and temporal aspects. Furthermore, the system categorizes the model's capabilities into different levels.

The multiple-choice task for IPEval is defined using the following three sections:

Question: The question is described in text form, which can be a short text directly asking about relevant knowledge or a longer text providing some context information.

Answer: Each question provides multiple answer options. For Chinese questions, there are 4 options (A, B, C, D), while English questions have 5 options (A, B, C, D, E). The answer to each question may be one or multiple options.

Explanation: The answer explanation provides further clarification for the question. Chinese questions do not have specific answer explanations, while English questions include answer explanations. These explanations can be used for understanding the train of thought and as references for further research.

With these sections, each question forms a triplet of (Question, Answer, Explanation). The goal of the task is to determine one or multiple correct answers for each question.

\section{Model being Evaluated}
\label{B}
In the evaluation of IPeval conducted in this study, as shown in Table, we assessed 14 models originating from different countries, with their primary training languages being English or Chinese, varying degrees of openness, different parameter sizes, and varying levels of generality. In the evaluation of closed-source commercial models, we primarily assessed models such as ChatGPT represented by GPT-3.5-turbo, GPT-4.0, GLM4, moonshot-v1-8k, and Qwen-max . Most of these closed-source models have parameters numbering in the hundreds of billions, yet the specific parameter counts have not been disclosed by the respective organizations. In the evaluation of open-source models, we primarily assessed ChatGLM3-6B, GLM-3-turbo , Baichuan2-7B/13B-Chat-v1 , Qwen-7B/72B-Chat, and Qwen1.5-14B-Chat. The parameter sizes of these open-source models, except for GLM-3-turbo, are all below 100B. The aforementioned closed-source and open-source models are all general-purpose large language models. In addition to general-purpose LLMs, given the intersection between the intellectual property and legal domains, we also evaluated a large model fine-tuned on legal corpora. This model, named "fuzi-mingcha-v1", was developed by the SDUIR . It is fine-tuned based on the ChatGLM-6B base model and claims to achieve superior performance compared to various other legal domain LLMs such as ChatLaw and LawGPT.

\textbf{GPT-3.5-turbo (ChatGPT) and GPT-4} are models from the GPT series developed by OpenAI. ChatGPT pioneers the development of large language models, both of which support efficient and fluent conversational modes. Moreover, GPT-4 is capable of performing visual multimodal dialogue tasks. Both models have undergone pre-training on large-scale corpora in multiple languages and have been further trained with techniques such as human value alignment and instruction following. They are capable of effectively executing human commands and providing safe and reliable results. GPT-4 also possesses the ability to read various file formats, thereby enhancing result accuracy by accessing third-party knowledge sources.

\textbf{Moonshot-v1-8k} is a text generation model provided by Moonshot, trained to understand natural language and written text. It is capable of generating text output based on input and can perform tasks such as content or code generation, summarization, dialogue, and creative writing. The "8k" in its model designation refers to the maximum context length of 8k, which includes both the input message and the generated output. Additionally, Moonshot also provides models with context lengths of 32k and 128k, which do not differ significantly in performance. The API provided by Moonshot's dark side does not support features such as web searching. However, its free online service, KimiChat, enables web searching to provide users with content from clear sources. Additionally, moonshot-v1-8k enhances its generation capabilities by allowing retrieval through the upload of documents in various formats.

\textbf{Qwen-7B-Chat and Qwen-72B-Chat} are large language models developed by Alibaba Cloud as part of the Qwen series. Both models are based on Transformer architecture and pretrained on data consisting of over 30 trillion tokens, encompassing high-quality data in Chinese, English, multilingual texts, code, mathematics, and other domains. Extensive comparative experiments were conducted to optimize the distribution of the pretraining corpus. In addition to diverse training data types covering a wide range, including a large amount of web text, professional books, and code, both datasets have undergone alignment mechanisms to adjust the initial Qwen-7B and Qwen-72B base models. Among them, Qwen-72B covers a more comprehensive vocabulary of approximately 150,000 words and is more multilingual-friendly, enabling users to enhance and expand capabilities in certain languages without expanding the vocabulary. Additionally, Qwen-72B supports a context length of 32k, enabling capabilities such as role-playing, language style transfer, task setting, and behavior setting through the adjustment of system commands.

\textbf{Qwen1.5-14B-Chat} is the beta version of Qwen2, developed by Alibaba Cloud. It is a decoder-based language model built upon Transformer architecture. Pre-training of the model is conducted using extensive data, followed by supervised fine-tuning and direct preference optimization during post-training. Qwen1.5-14B-Chat inherits the advantages of its predecessor versions in the Qwen series and features enhancements such as SwiGLU activation, attention QKV bias, group query attention, sliding window attention, and a mixture of full attention. We have also introduced an enhanced tokenizer capable of adapting to multiple natural languages and code, significantly improving human conversational preferences and multilingual support capabilities, while stably supporting a context length of up to 32k.

\textbf{Qwen-Max} is a billion-scale ultra-large language model from the Qwen series, supporting input in different languages such as Chinese and English, with a context length of up to 8k tokens. It has been pre-trained on more extensive corpora and demonstrates outstanding performance across various downstream tasks in both Chinese and English.

\textbf{Baichuan2-7B-Chat and Baichuan2-13B-Chat} are the next-generation open-source large language models introduced by Baichuan Intelligence. Based on the Transformer architecture, they are trained on high-quality corpora consisting of 26 trillion tokens and support both Chinese and English languages, with a context window length of 4096. They achieve state-of-the-art performance among models of similar sizes across various authoritative benchmarks in Chinese, English, and multilingual general and domain-specific domains.

\textbf{ChatGLM3-6B, GLM-3-turbo, and GLM-4} are dialogue pre-training models jointly released by Zhifu AI and the KEG Laboratory of Tsinghua University. Among them, ChatGLM3-6B is an open-source model in the ChatGLM3 series, which inherits numerous outstanding features such as smooth dialogue flow and low deployment threshold from the previous two generations of models, while also possessing many new characteristics. These include the utilization of a more diverse training dataset, more extensive training steps, and a more rational training strategy. Evaluation on datasets from various perspectives including semantics, mathematics, reasoning, code, and knowledge shows that ChatGLM3-6B-Base exhibits the strongest performance among base models with fewer than 10B parameters. The overall performance of the GLM-4 model has significantly improved compared to the previous generation, with over a dozen metrics approaching or reaching the level of GPT-4. It supports longer contexts, stronger multimodality, faster inference speed, increased concurrency, and significantly reduced inference costs. Additionally, GLM-4 enhances the capabilities of intelligent agents.

\textbf{Fuzi-mingcha-v1*} is a Chinese judicial large-scale model jointly developed by Shandong University, Inspur Cloud, and China University of Political Science and Law. It is built upon ChatGLM as the large model base and trained on a massive Chinese corpus of unsupervised judicial data, including various types of judgments, laws, and regulations, as well as supervised judicial fine-tuning data, including legal Q\&A and case retrieval. The model supports functions such as legal provision retrieval, case analysis, syllogistic reasoning judgment, and judicial dialogue, aiming to provide users with comprehensive and highly accurate legal consulting and answer services. It features three main characteristics: provision-based retrieval response, case-based retrieval response, and syllogistic reasoning judgment. It can generate responses based on relevant legal provisions, analyze input cases based on historically similar cases, and automatically analyze the case details to identify key facts and legal regulations, thereby generating a logically rigorous syllogistic judgment prediction for specific cases. In September 2023, the LawBench judicial competency evaluation system, jointly launched by the Shanghai AI Lab and Nanjing University, excelled in the zero-shot performance of its law-specific LLMs, achieving first place. This performance marked a significant improvement compared to ChatGLM, which had not been trained on legal expertise.

\textbf{MoZi-bloomz*} is a vertical large model specifically designed for the field of intellectual property, based on the BLOOMZ multilingual large-scale language model. It underwent pre-training, instruction fine-tuning, and IP-specific instruction fine-tuning. MoZi-bloomz was pre-trained using 23 million official patent documents and fully fine-tuned with instructions. MoZi-bloomz excels in the intellectual property domain, including tasks such as IP multiple-choice questions, IP question answering, and patent matching. Experiments show that MoZi-bloomz outperforms models such as BLOOMZ, BELLE, and ChatGLM in the IP field.

\textbf{MoZi-glm*} is fine-tuned based on ChatGLM and trained with a large amount of Chinese legal texts, including judgments, laws, regulations, and legal Q\&A. MoZi-glm supports functions such as legal clause retrieval, case analysis, syllogistic reasoning judgment, and judicial dialogue. This model performed excellently in the LawBench judicial capability evaluation system, especially in the zero-shot performance for legal tasks, showing significant improvement compared to ChatGLM.

\section{More Results}
For more results, please see table\ref{tab:table7}, \ref{tab:table8}, \ref{tab:table9}, \ref{tab:table10}.
\label{C}
\begin{figure}[htbp]
	\centering
	\includegraphics[width = \linewidth]{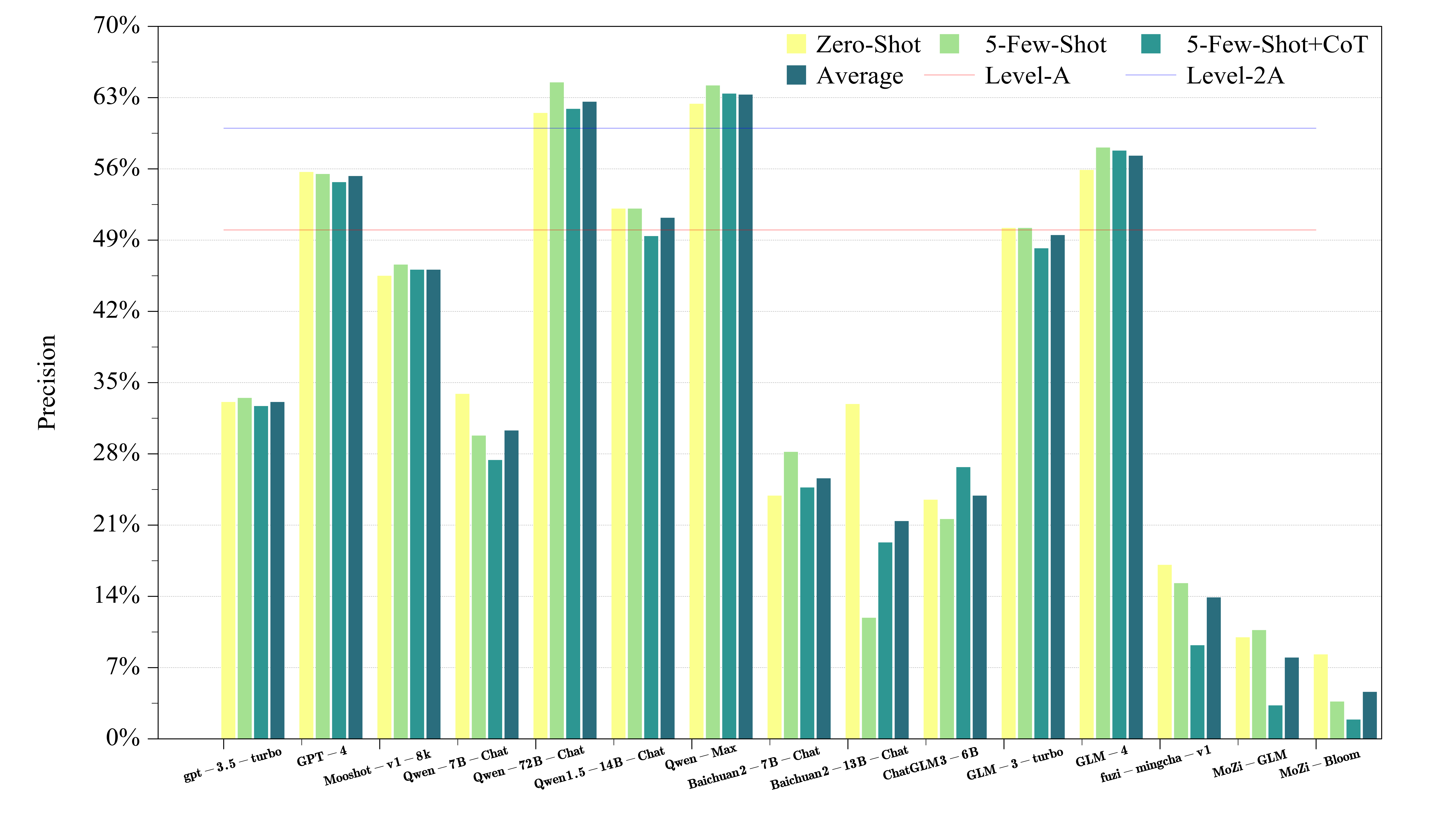}
	\caption{Main result. We evaluated fifteen models using three prompt strategies.}
	\label{main}
\end{figure}

\begin{table}[htbp]
	\centering
	\caption{Different language test results. We conducted performance statistics of the models on both Chinese and English data in IPEval, employing three prompt strategies.}
	\resizebox{13.5cm}{!}{
		\begin{tblr}{
				cells = {c},
				cell{1}{1} = {r=2}{},
				cell{1}{2} = {c=2}{},
				cell{1}{4} = {c=2}{},
				cell{1}{6} = {c=2}{},
				vline{3,5} = {1}{},
				vline{4,6} = {2-17}{},
				hline{1,18} = {-}{0.08em},
				hline{2} = {2-7}{},
				hline{3,5-6,10,12,15} = {-}{},
			}
			\textbf{Model}                     & \textbf{Zero-Shot} &                  & \textbf{5-Few-Shot} &                  & \textbf{5-Few-Shot+CoT} &                  \\
			& \textbf{Chinese}   & \textbf{English} & \textbf{Chinese}    & \textbf{English} & \textbf{Chinese}        & \textbf{English} \\
			gpt-3.5-turbo(ChatGPT)             & 29.0               & 38.5             & 28.0                & 40.5             & 29.0                    & 37.4             \\
			GPT-4                              & 53.4               & \textbf{58.6}    & 52.8                & \textbf{59.0}    & 51.6                    & \textbf{58.6}    \\
			moonshot-v1-8k                     & 47.9               & 42.4             & 49.3                & 43.2             & 50.4                    & 40.5             \\
			Qwen-7B-Chat                       & 35.9               & 31.2             & 29.3                & 30.4             & 28.4                    & 26.1             \\
			Qwen-72B-Chat                      & \textbf{70.7}      & 49.7             & \textbf{77.3}       & 48.2             & 76.2                    & 43.6             \\
			Qwen1.5-14B-Chat                   & 60.5               & 41.3             & 59.5                & 42.7             & 58.6                    & 37.6             \\
			Qwen-Max                           & 70.6               & 52.0             & 73.6                & 52.2             & \textbf{77.2}           & 45.7             \\
			Baichuan2-7B-Chat                  & 18.9               & 30.4             & 26.2                & 30.8             & 23.9                    & 25.6             \\
			Baichuan2-13B-Chat                 & 32.0               & 33.9             & 10.0                & 14.4             & 21.8                    & 16.0             \\
			ChatGLM3-6B                        & 18.3               & 30.1             & 21.5                & 21.8             & 25.3                    & 28.5             \\
			GLM-3-turbo                        & 56.8               & 41.8             & 57.5                & 40.9             & 57.1                    & 36.8             \\
			GLM-4                              & 62.8               & 47.1             & 64.6                & 49.8             & 64.8                    & 48.8             \\
			fuzi-mingcha-v1\textsuperscript{*} & 13.7               & 21.3             & 13.0                & 18.2             & 9.2                     & -\\                
			MoZi-glm\textsuperscript{*}	&	6.0		&	17.9	&	6.3		&	18.6	&	3.3	 	&	-	\\
			MoZi-bloomz\textsuperscript{*}	&	12.0	&	0.7		&		3.7		&	3.4		&1.9	&	-
			
		\end{tblr}
	}
	\label{tab:table7}
\end{table}

\begin{figure}[htbp]
	\centering
	\includegraphics[width = \linewidth]{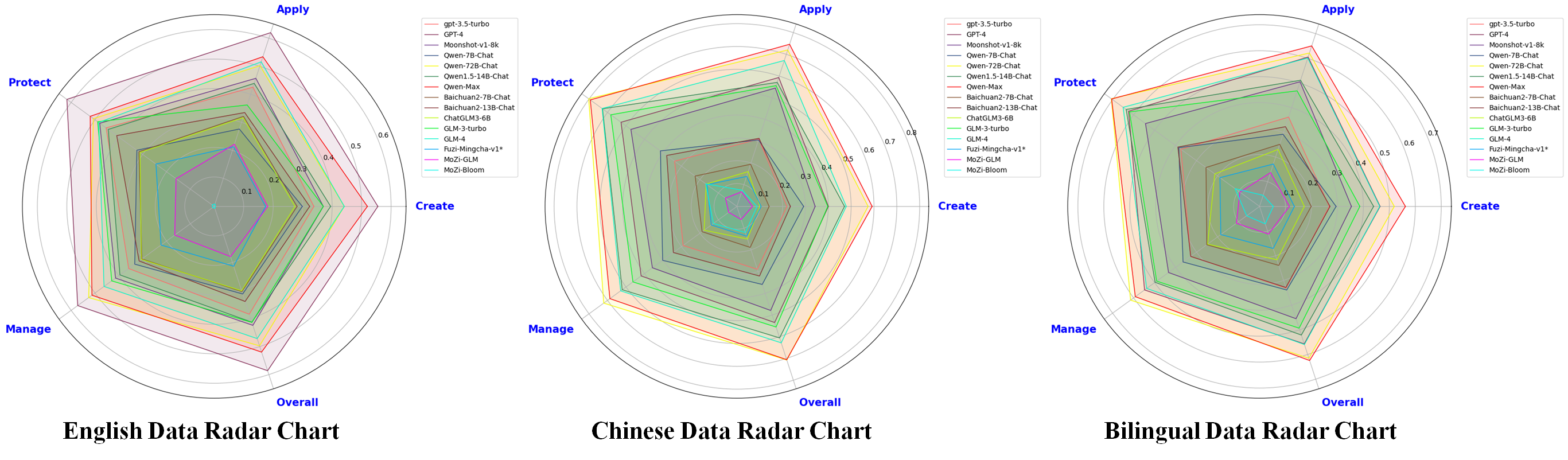}
	\caption{Model ability radar chart. We evaluated the performance of fifteen models from five perspectives: creation, application, protection, management, and overall performance. Among them, the Qwen and GPT series models performed well.}
	\label{radar}
\end{figure}

\begin{table}[htbp]
	\centering
	\caption{Results of different question types testing. We conducted statistical analysis and evaluation of the models' performance on single-choice and multiple-choice questions.}
	\resizebox{13.5cm}{!}{
		\begin{tblr}{
				cells = {c},
				cell{1}{1} = {r=2}{},
				cell{1}{2} = {c=2}{},
				cell{1}{4} = {c=2}{},
				cell{1}{6} = {c=2}{},
				vline{3,5} = {1}{},
				vline{4,6} = {2-17}{},
				hline{1,18} = {-}{0.08em},
				hline{2} = {2-7}{},
				hline{3,5-6,10,12,15} = {-}{},
			}
			\textbf{Model}                     & \textbf{Zero-Shot} &                   & \textbf{5-Few-Shot} &                   & \textbf{5-Few-Shot+CoT} &                   \\
			& \textbf{Single}    & \textbf{Multiple} & \textbf{Single}     & \textbf{Multiple} & \textbf{Single}         & \textbf{Multiple} \\
			gpt-3.5-turbo(ChatGPT)             & 42.8               & 23.0              & 42.6                & 21.7              & 35.7                    & 26.1              \\
			GPT-4                              & 67.4               & 47.4              & 65.4                & 47.3              & 62.1                    & 48.1              \\
			moonshot-v1-8k                     & 62.7               & 41.5              & 57.0                & 46.0              & 59.6                    & 46.4              \\
			Qwen-7B-Chat                       & 50.3               & 29.7              & 48.1                & 21.2              & 39.9                    & 23.4              \\
			Qwen-72B-Chat                      & \textbf{82.3}      & 65.7              & \textbf{86.9}       & \textbf{73.1}     & 86.9                    & 71.6              \\
			Qwen1.5-14B-Chat                   & 73.6               & 54.9              & 70.5                & 63.2              & 73.2                    & 52.3              \\
			Qwen-Max                           & 79.6               & \textbf{66.7}     & 79.4                & 71.2              & \textbf{87.7}           & \textbf{72.8}     \\
			Baichuan2-7B-Chat                  & 42.4               & 8.7               & 43.7                & 18.7              & 45.2                    & 14.8              \\
			Baichuan2-13B-Chat                 & 47.0               & 25.6              & 28.8                & 1.9               & 28.2                    & 19.1              \\
			ChatGLM3-6B                        & 38.6               & 9.5               & 41.5                & 12.9              & 44.3                    & 17.1              \\
			GLM-3-turbo                        & 70.7               & 51.3              & 64.3                & 54.6              & 63.9                    & 54.1              \\
			GLM-4                              & 74.9               & 57.6              & 73.2                & 61.4              & 78.0                    & 59.1              \\
			fuzi-mingcha-v1\textsuperscript{*} & 30.8               & 6.3               & 26.6                & 7.1               & 17.9                    & 5.4               \\
			MoZi-glm\textsuperscript{*}		&		15.9		&		4.4		&	14.3		&		8.5		&		4.9		&		2.6\\		
			MoZi-bloomz\textsuperscript{*}		&		4.4		&		11.0		&		4.1		&		3.0		&		1.8		&	1.9
		\end{tblr}
	}
	\label{tab:table8}
\end{table}

\begin{table}[htbp]
	\centering
	\caption{Results of testing patent law and relevant laws in Chinese questions. We evaluated and analyzed the models' performance from the perspectives of patent law and related laws on Chinese data.}
	\resizebox{13.5cm}{!}{
		\begin{tblr}{
				cells = {c},
				cell{1}{1} = {r=2}{},
				cell{1}{2} = {c=2}{},
				cell{1}{4} = {c=2}{},
				cell{1}{6} = {c=2}{},
				vline{3,5} = {1}{},
				vline{4,6} = {2-17}{},
				hline{1,18} = {-}{0.08em},
				hline{2} = {2-7}{},
				hline{3,5-6,10,12,15} = {-}{},
			}
			\textbf{Model}                     & \textbf{Zero-Shot} &                    & \textbf{5-Few-Shot} &                    & \textbf{5-Few-Shot+CoT} &                    \\
			& \textbf{Patent law}       & \textbf{Relevant law}       & \textbf{Patent law}        & \textbf{Relevant law}       & \textbf{Patent law}            & \textbf{Relevant law}       \\
			gpt-3.5-turbo(ChatGPT)             & 35.5/16.8          & 51.2/30.1          & 38.4/14.6           & 47.4/29.9          & 25.6/20.0               & 47.4/32.9          \\
			GPT-4                              & 60.3/33.7          & 75.6/63.0          & 57.4/34.1           & 74.6/62.4          & 56.2/34.4               & 68.9/63.6          \\
			moonshot-v1-8k                     & 51.2/28.5          & 76.1/56.2          & 46.7/30.5           & 68.9/63.6          & 48.8/31.0               & 72.2/63.8          \\
			Qwen-7B-Chat                       & 40.5/20.5          & 61.7/40.1          & 38.0/10.8           & 59.8/32.9          & 30.6/13.3               & 50.7/34.8          \\
			Qwen-72B-Chat                      & \textbf{76.0/54.8} & \textbf{89.5}/78.1 & \textbf{83.1/63.6}  & \textbf{91.4}/83.8 & 83.1/59.6               & \textbf{91.4/85.1} \\
			Qwen1.5-14B-Chat                   & 63.6/40.7          & 85.2/71.0          & 59.5/56.0           & 83.3/71.4          & 64.5/37.1               & 83.3/69.5          \\
			Qwen-Max                           & 73.1/53.7          & 87.1/\textbf{81.4} & 72.7/59.5           & 87.1/\textbf{84.5} & \textbf{86.1/62.4}      & 89.5/84.7          \\
			Baichuan2-7B-Chat                  & 30.2/5.2           & 56.5/12.7          & 36.0/12.4           & 52.6/25.8          & 33.5/8.8                & 58.9/21.5          \\
			Baichuan2-13B-Chat                 & 38.4/18.7          & 56.9/33.3          & 25.6/1.4            & 32.5/2.5           & 21.9/16.2               & 35.4/22.3          \\
			ChatGLM3-6B                        & 32.6/6.5           & 45.5/12.9          & 34.7/11.0           & 49.3/15.1          & 36.8/10.5               & 53.1/24.7          \\
			GLM-3-turbo                        & 57.4/35.6          & 86.1/69.1          & 50.4/37.5           & 80.4/74.0          & 51.7/37.3               & 78.0/73.2          \\
			GLM-4                              & 64.0/42.9          & 87.6/74.2          & 62.8/45.0           & 85.2/79.9          & 69.8/43.2               & 87.6/77.1          \\
			fuzi-mingcha-v1\textsuperscript{*} & 21.9/5.6           & 41.1/7.2           & 25.2/5.6            & 28.2/8.8           & 21.9/4.7                & 13.2/6.3       \\
			MoZi-glm\textsuperscript{*} & 11.2/3.8		&		8.1/5.1		&		1.2/9.4	&3.3/8.0 & 2.9/3.2 &7.2/1.8	\\
			MoZi-bloomz\textsuperscript{*} &	10.7/11.2		&		16.7/11.7		&		5.4/2.3	&5.7/3.7	&2.9/0.9	&0.5/3.1
		\end{tblr}
	}
	\label{tab:table9}
\end{table}

\begin{table}[htbp]
	\centering
	\caption{Performance of models across different years. Each number represents the performance of the model on a specific year's exam questions, using the zero-shot strategy.}
	\resizebox{13.5cm}{!}{
		\begin{tblr}{
				cells = {c},
				cell{1}{1} = {r=2}{},
				cell{1}{2} = {c=4}{},
				cell{1}{6} = {c=4}{},
				vline{3} = {1}{},
				vline{6} = {2-17}{},
				hline{1,18} = {-}{0.08em},
				hline{2} = {2-9}{},
				hline{3,5-6,10,12,15} = {-}{},
			}
			\textbf{Model}                     & \textbf{Chinese(Patent law/Relevant law)}   &                    &                    &                    & \textbf{English}   &               &                    &                    \\
			& \textbf{2012-2013} & \textbf{2014-2015} & \textbf{2016-2017} & \textbf{2018-2019} & \textbf{1997-1998} & \textbf{1999} & \textbf{2000-2001} & \textbf{2002-2003} \\
			gpt-3.5-turbo(ChatGPT)             & 25.0/36.4          & 20.6/37.2          & 19.1/44.0          & 25.1/28.0          & 35.2               & 33.3          & 37.8               & 43.1               \\
			GPT-4                              & 45.5/72.7          & 42.2/70.4          & 37.2/71.0          & 42.2/55.5          & \textbf{47.3}      & \textbf{56.6} & \textbf{58.0}      & \textbf{64.9}      \\
			moonshot-v1-8k                     & 47.0/65.7          & 28.6/62.3          & 29.2/70.0          & 36.7/52.5          & 41.2               & 37.9          & 42.5               & 45.1               \\
			Qwen-7B-Chat                       & 31.5/49.5          & 22.6/46.7          & 23.6/54.5          & 28.6/37.0          & 32.1               & 30.3          & 31.5               & 31.1               \\
			Qwen-72B-Chat                      & \textbf{65.0}/84.9 & 62.3/81.4          & \textbf{88.9/91.0} & \textbf{58.8}/70.5 & \textbf{47.3}      & 45.0          & 49.8               & 53.1               \\
			Qwen1.5-14B-Chat                   & 46.5/77.8          & 50.8/77.9          & 45.7/81.0          & 47.7/65.5          & 38.2               & 38.9          & 40.0               & 45.1               \\
			Qwen-Max                           & 60.0/\textbf{87.9} & \textbf{63.8/83.9} & 60.8/90.5          & 53.8/72.5          & 46.1               & 52.5          & 51.8               & 54.4               \\
			Baichuan2-7B-Chat                  & 14.5/28.3          & 13.6/24.1          & 10.6/34.0          & 12.6/18.0          & 22.4               & 26.3          & 34.3               & 31.8               \\
			Baichuan2-13B-Chat                 & 31.0/42.4          & 20.6/39.2          & 21.6/48.0          & 25.6/33.0          & 39.4               & 32.3          & 31.3               & 35.1               \\
			ChatGLM3-6B                        & 10.5/17.2          & 11.6/19.1          & 11.1/20.0          & 16.6/15.5          & 28.5               & 27.8          & 32.8               & 29.3               \\
			GLM-3-turbo                        & 43.0/79.8          & 37.2/74.4          & 42.2/84.0          & 43.7/52.0          & 40.0               & 34.3          & 42.0               & 44.9               \\
			GLM-4                              & 55.0/84.9          & 48.7/82.9          & 43.7/83.5          & 49.8/65.0          & 42.4               & 42.9          & 49.5               & 51.4               \\
			fuzi-mingcha-v1\textsuperscript{*} & 12.5/16.2          & 10.6/19.1          & 9.1/22.0           & 10.1/11.5          & 26.1               & 17.2          & 19.8               & 23.1          \\
			MoZi-glm\textsuperscript{*} & 5.6/7.1          & 4.5/6.0          & 7.5/9.0           & 6.5/2.5          & 22.1               &    19.5       & 16.4               & 16.5               \\
			MoZi-bloomz\textsuperscript{*} & 15.5/15.2          & 7.0/14.6          & 8.0/15.5           & 13.6/8.5          & 0.1               & 0.1          & 1.9               & 0.3          
		\end{tblr}
	}
	\label{tab:table10}
\end{table}

\begin{figure}[htbp]
	\centering
	\includegraphics[width = \linewidth]{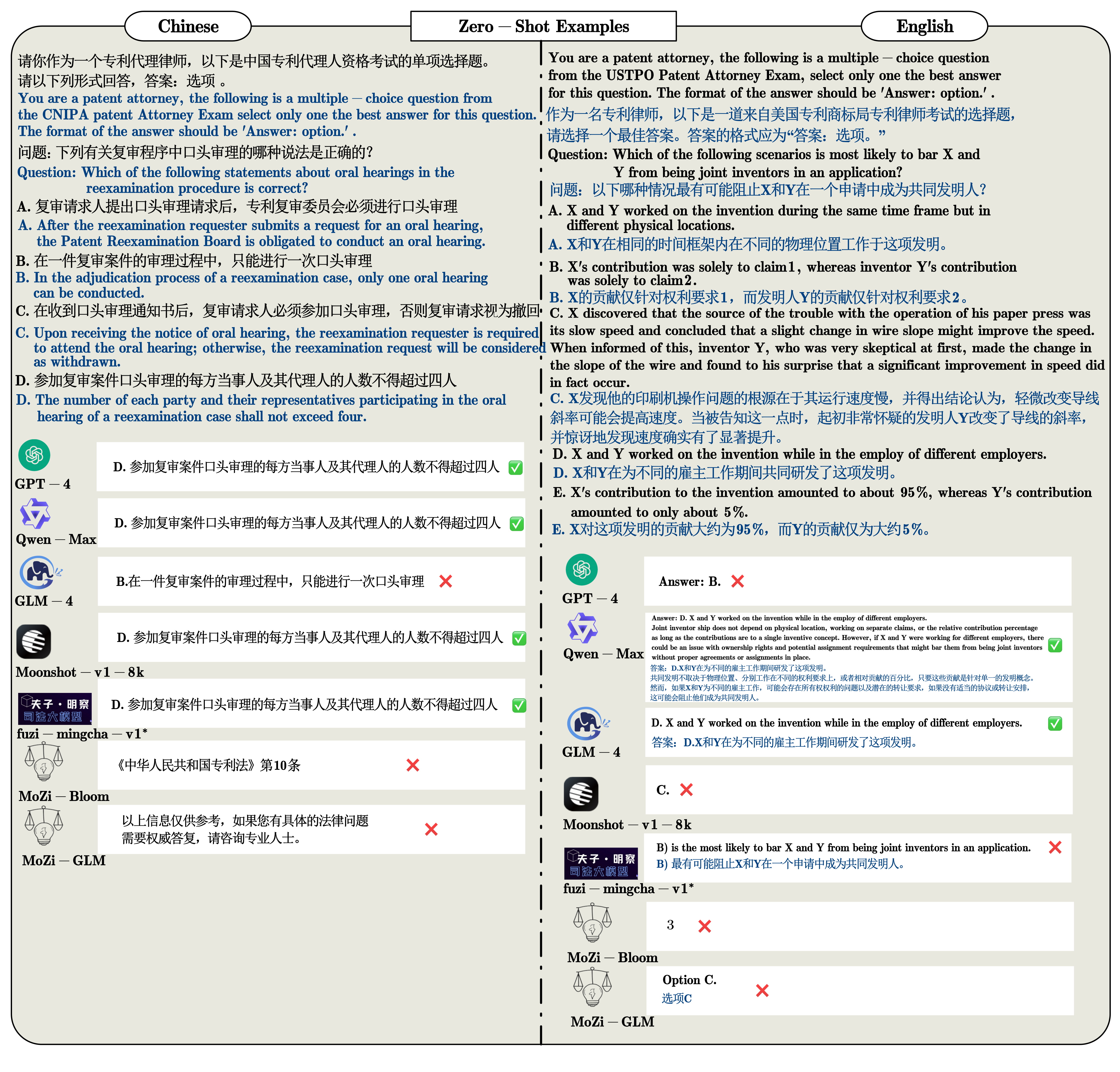}
	\caption{Zero-shot examples. On both Chinese and English data, the performance of the same model can be observed, indicating that the performance of the model on IPEval is not satisfactory.}
	\label{zeroshot}
\end{figure}

\begin{figure}
	\centering
	\includegraphics[width = \linewidth]{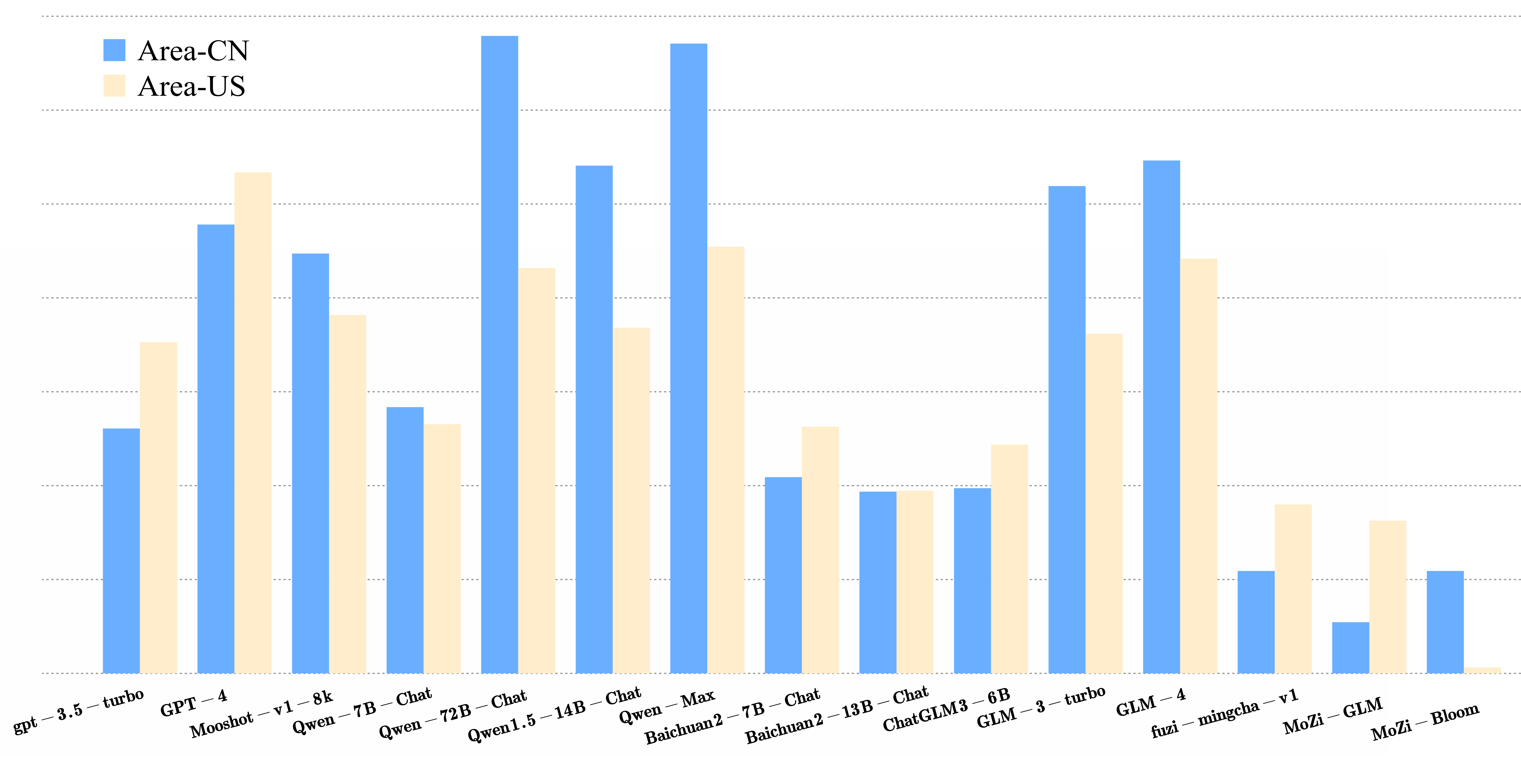}
	\caption{Model performance in different area. The performance of the model on Chinese and English data demonstrates the regional characteristics of the dataset. For more analysis, please refer to the Appendix~\ref{D.3}.}
	\label{Area}
\end{figure}

\begin{figure}
	\centering
	\includegraphics[width = \linewidth]{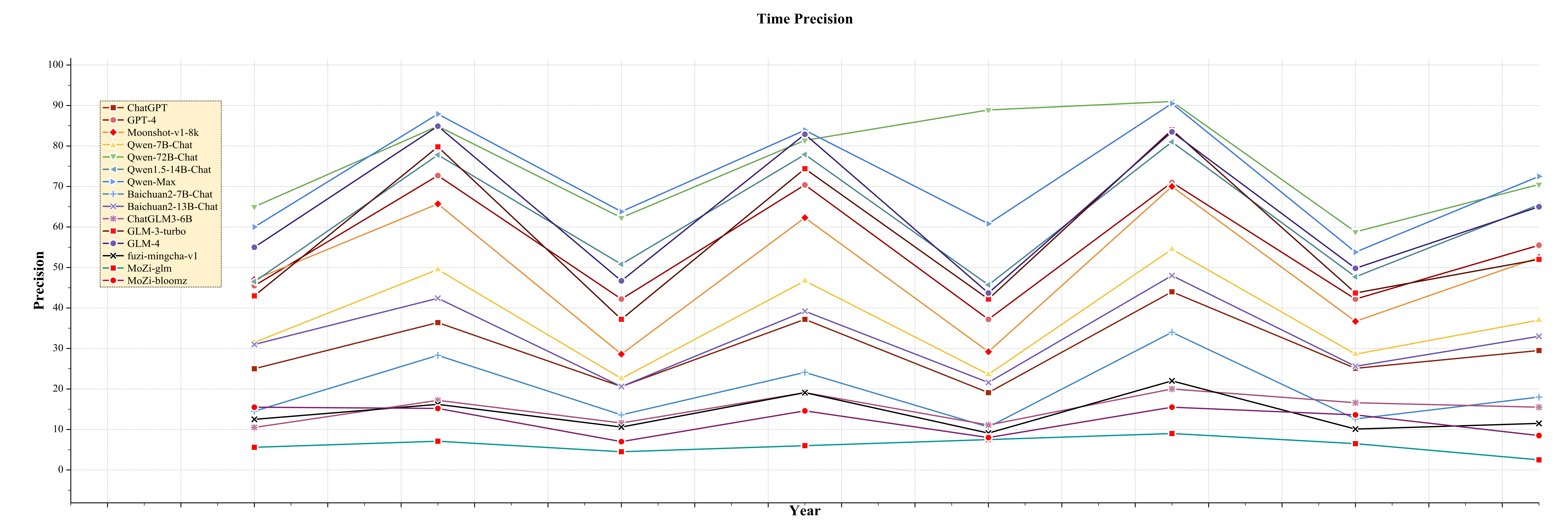}
	\caption{Performance of the model on Chinese data over the years. The temporality of the dataset can be observed. For further analysis, please refer to the Appendix~\ref{D.3}.}
	\label{Chinese_time}
\end{figure}

\begin{figure}
	\centering
	\includegraphics[width = \linewidth]{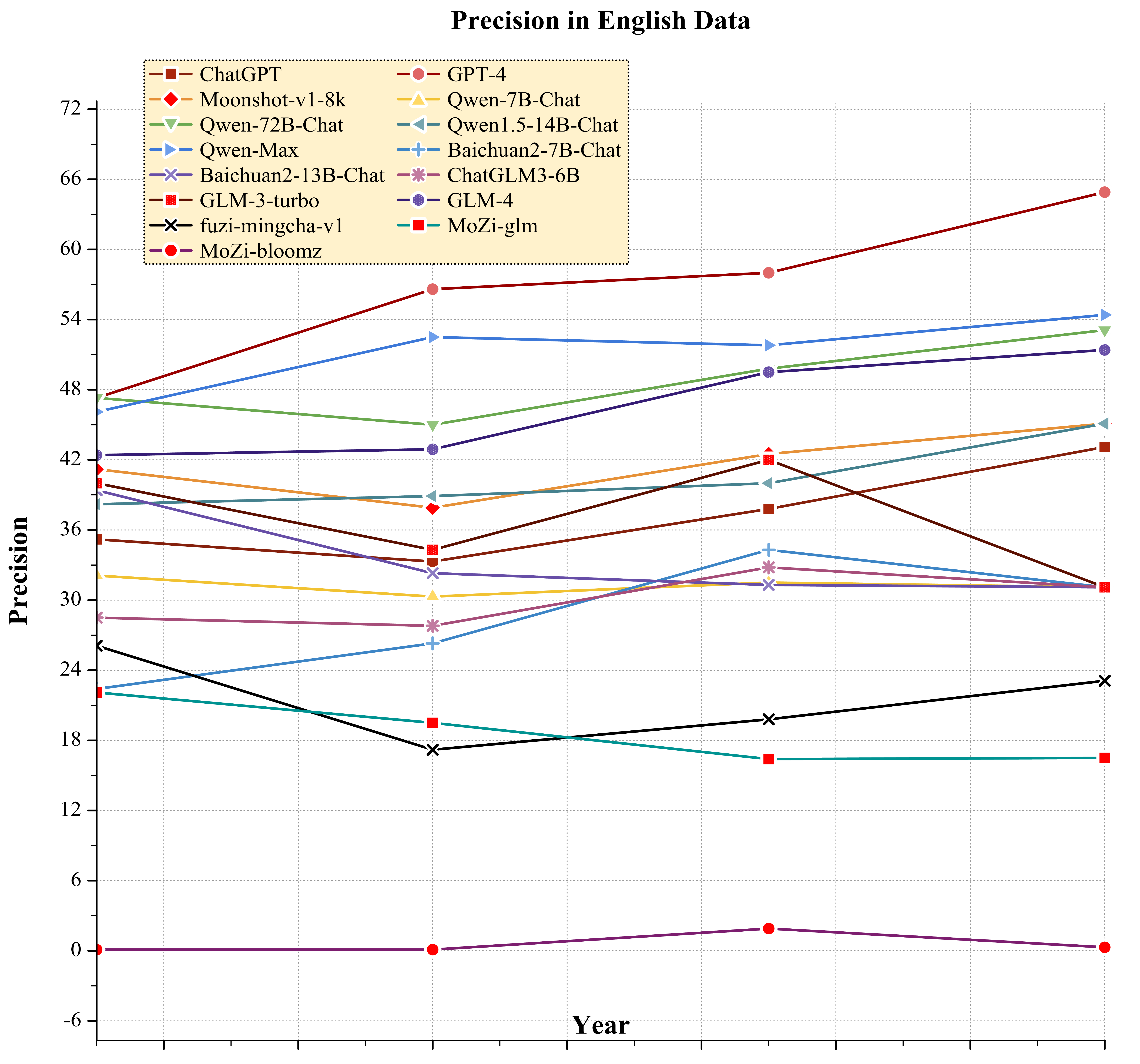}
	\caption{Performance of the model on English data over the years. The temporality of the dataset can be observed. For further analysis, please refer to the Appendix~\ref{D.3}.}
	\label{English_time}
\end{figure}

\section{Analysis}
\label{D}
\subsection{Performance Analysis}
\label{D.1}

\begin{figure}[htbp]
	\centering
	\includegraphics[width = \linewidth]{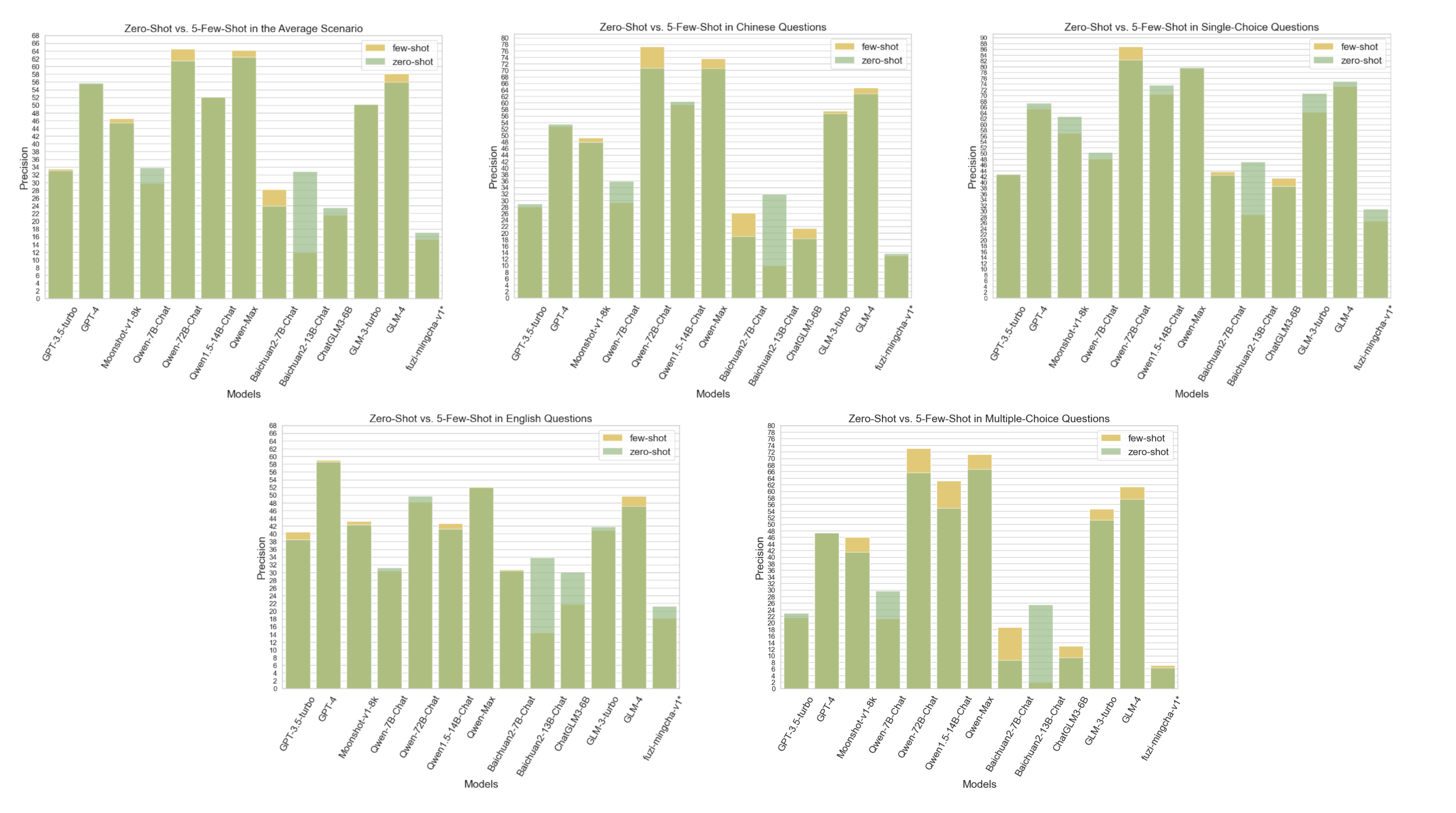}
	\caption{Zero-Shot vs.5-Few-Shot. It can be observed that the few-shot strategy did not significantly improve performance.}
	\label{zf}
\end{figure}

\begin{figure}[htbp]
	\centering
	\includegraphics[width = \linewidth]{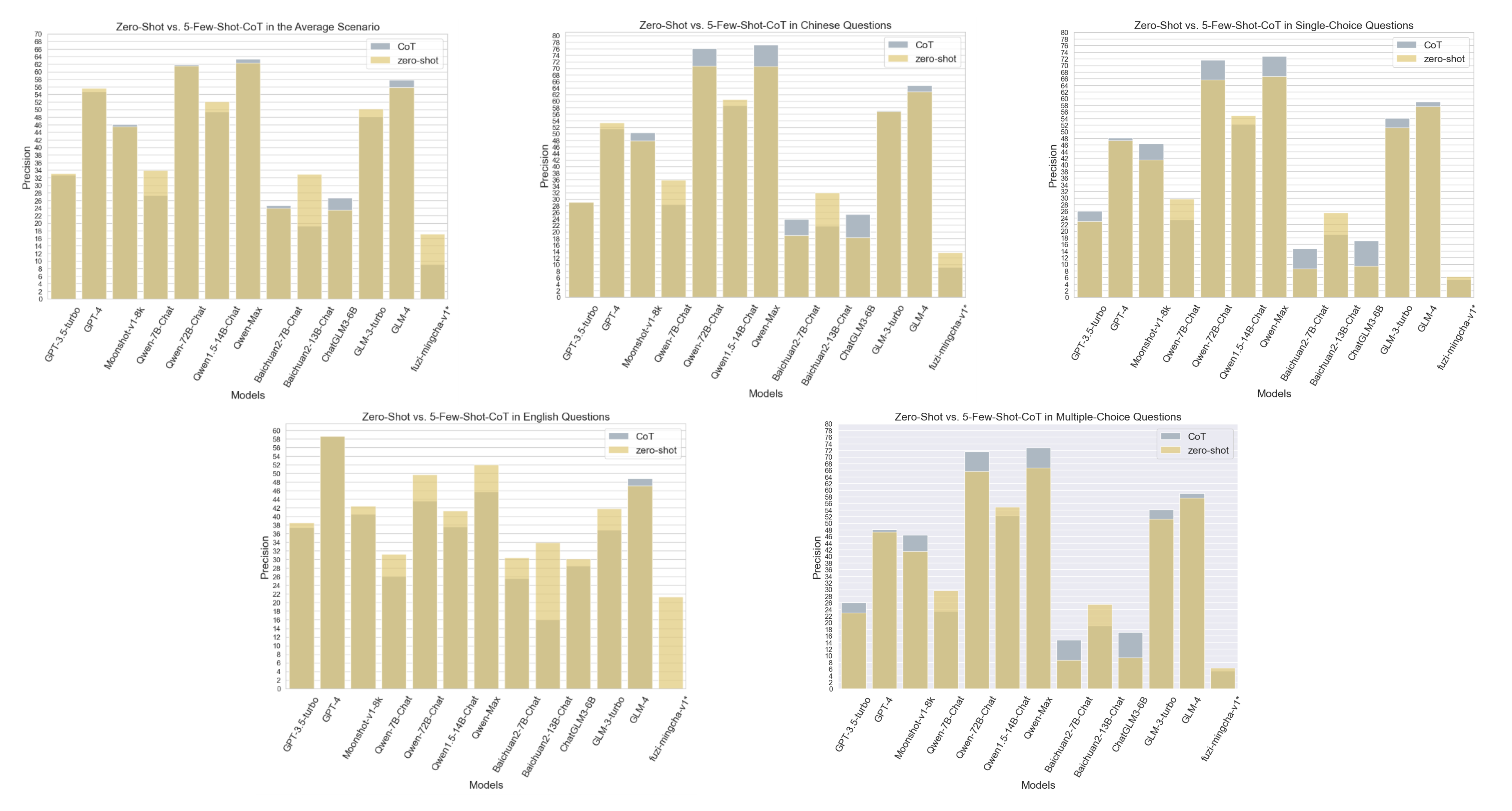}
	\caption{Zero-Shot vs.5-Few-Shot-CoT. It can be observed that, compared to the zero-shot strategy, the CoT strategy also did not significantly improve performance.}
	\label{zc}
\end{figure}

\begin{figure}
	\centering
	\includegraphics[width=\linewidth]{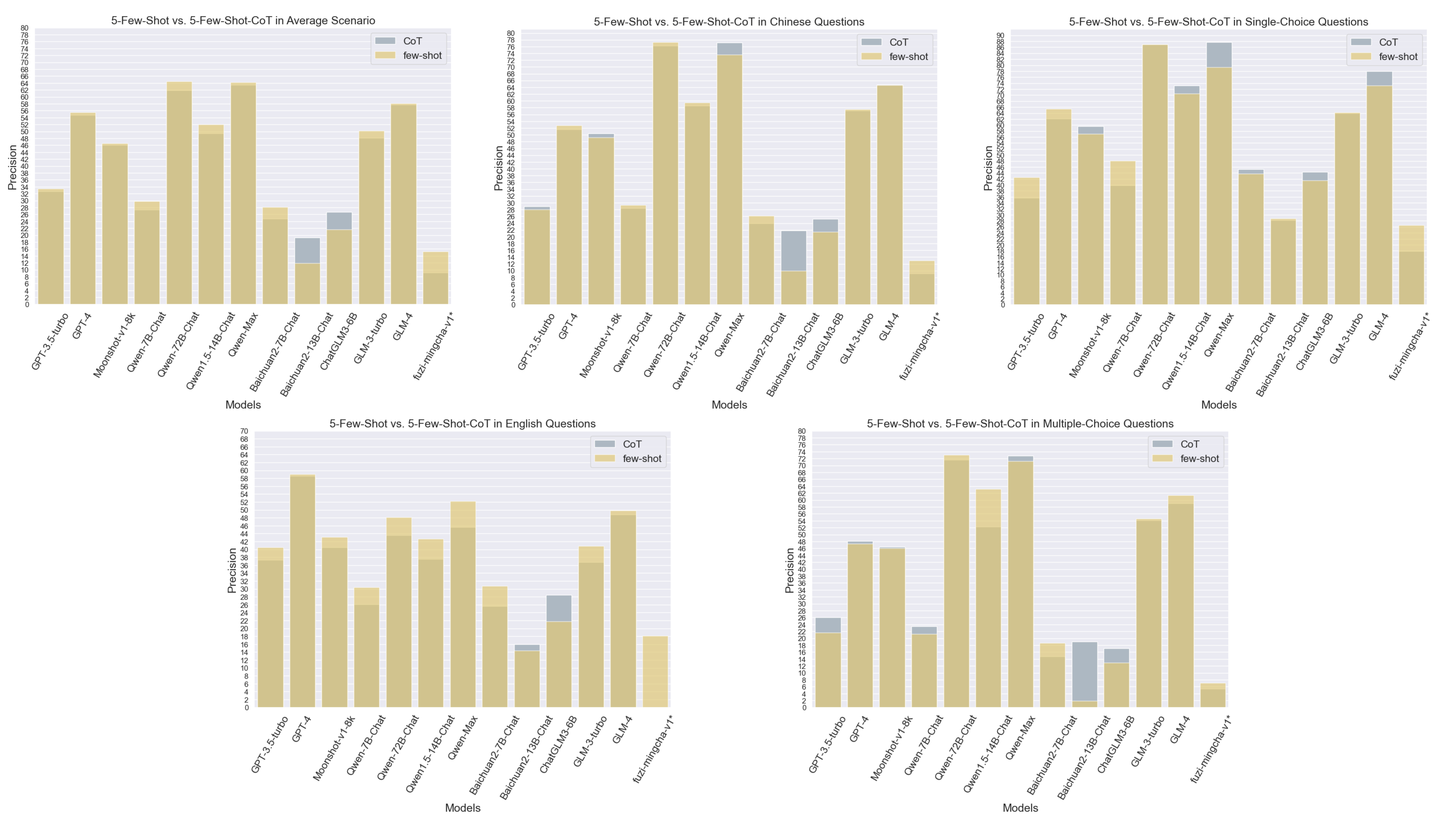}
	\caption{5-Few-Shot vs.5-Few-Shot-CoT. Most models perform better with the few-shot strategy than with the CoT strategy.}
	\label{fc}
\end{figure}

\paragraph{Is the model better at answering intellectual property questions in English or Chinese?}

The difference between Chinese and English data ensures the regional locality of the data. Furthermore, upon further observation, it can be noted that the GPT series, which is primarily trained in English, outperforms other models of similar parameter levels on English data. Similarly, the Qwen series models, primarily trained in Chinese, outperform other models of similar parameter levels on Chinese data. This indicates that the model's training corpus leads to a regional preference in intellectual property consulting tasks, consistent with our previous analysis.

We hope this regional characteristic can inspire relevant work: when applying large-scale models in the field of intellectual property, it is necessary to consider the regional locality of relevant knowledge and achieve balance in pre-training corpora. This characteristic also presents new challenges for future research: how to guide the model to understand the regional differences in intellectual property consulting tasks?

\paragraph{Is the model better at single-choice questions or multiple-choice questions?}

Upon further observation of Table \ref{tab:table8}, we found that in the Chinese test, all models performed significantly better on single-choice questions than on multiple-choice questions. This is particularly evident in smaller parameter models, with some models' accuracy on multiple-choice questions not even reaching 10\% (Baichuan2-7B-Chat, Fuzi-mingcha-v1*). This is because multiple-choice questions are much more difficult than single-choice questions, requiring stronger reasoning abilities from the large model. Compared to single-choice questions, multiple-choice questions not only require the model to accurately determine the correctness of each option but also to discern the relationships between options and between the question stem and the options. Smaller parameter models, due to their relatively limited domain knowledge and weaker reasoning capabilities, show a significant performance difference between single-choice and multiple-choice questions. Future research should focus on how to enhance the reasoning capabilities of smaller models in the field of intellectual property.

\paragraph{Did Few-Shot positively guide the model's performance?}
In Tables \ref{tab:table7}, \ref{tab:table8}, and \ref{tab:table9}, we have compiled the experimental results of 5-Few-Shot from different perspectives. The number of Few-Shots, treated as a hyperparameter, was determined to be 5 through practical experimentation. Most models showed a significant performance improvement after adopting the 5-Few-Shot strategy. For example, Baichuan2-7B-Chat's performance increased by 38.6\% on Chinese questions, and Qwen-72B-Chat's performance increased by 9.3\% on Chinese questions, reaching a highest score of 77.3, surpassing the larger Qwen-Max model.

For some models, the introduction of this strategy resulted in a steep performance decline. For instance, Baichuan2-13B-Chat experienced an average performance drop of 63.8\% after introducing this mode, even falling below its smaller counterpart, Baichuan2-7B-Chat. This may be due to the model framework and training mechanism being incompatible with Few-Shot learning. The Few-Shot strategy itself is not always effective: 1) While it can stimulate the model's knowledge in a specific domain to some extent and influence the distribution of model outputs, its main role is in standardizing the model's output format. 2) There is no guarantee that the randomly selected 5 examples are relevant to the question, nor can it ensure that there is no information leakage in the examples. Future research should focus on creating a higher-quality pool of intellectual property Few-Shot examples.

\paragraph{Did the chain of thought positively guide the model's performance?}
In Tables \ref{tab:table7}, \ref{tab:table8}, and \ref{tab:table9}, we have compiled the experimental results of the chain of thought from different perspectives. Almost all models experienced a performance decline when the chain of thought was introduced on top of the 5-Few-Shot strategy, except for ChatGLM3-6B, which saw a performance increase of 23.6\%. Baichuan2-13B-Chat's performance decreased by 41.3\% with the chain of thought compared to Zero-Shot, but it achieved a 62.2\% performance gain compared to Few-Shot (this model performed better in Zero-Shot than Few-Shot). Conversely, larger models such as GLM-4, Qwen-Max, and Moonshot-v1-8K all showed performance improvements with the chain of thought compared to Zero-Shot
.
With the introduction of the chain of thought, the length of the model's output significantly increased compared to Zero-Shot and Few-Shot, and it became more logically structured. Most models showed performance improvements on multiple-choice questions, as these questions require stronger reasoning abilities, and the chain of thought is a good strategy to enhance the model's reasoning capabilities. Under the chain of thought mode, Qwen-Max achieved the highest scores of 87.7 and 72.8 in Chinese single-choice and multiple-choice questions, respectively. As shown in Figures \ref{zc}, \ref{zf}, and \ref{fc}, we visualized the performance changes of the chain of thought compared to other modes in various scenarios using bar charts.

\paragraph{
	Can the model clearly differentiate between intellectual property laws from different times?}
Upon further observation of Table \ref{tab:table10} in Appendix \ref{C}, it can be seen that different series of models perform differently across different time periods. The GPT series performs better on Chinese questions from 2012-2013 and English questions from 2002-2003, while the Qwen series performs better on Chinese questions from 2016-2017 and English questions from 2002-2003. This may be related to the distribution differences of intellectual property data from different years in the model's training corpus. Intellectual property is a dynamic and evolving field with constantly changing and expanding knowledge. Although its fundamental core may not undergo significant changes, in order to better fulfill intellectual property consulting tasks, models should be able to grasp the differences in intellectual property knowledge across different periods, including the constantly updated laws. This requires researchers to construct relevant corpora from different time periods and enable models to perceive the differences in knowledge across different periods.

\subsection{Error Analysis}
\label{D.2}

\begin{figure}[htbp]
	\centering
	\includegraphics[width = 0.7\linewidth]{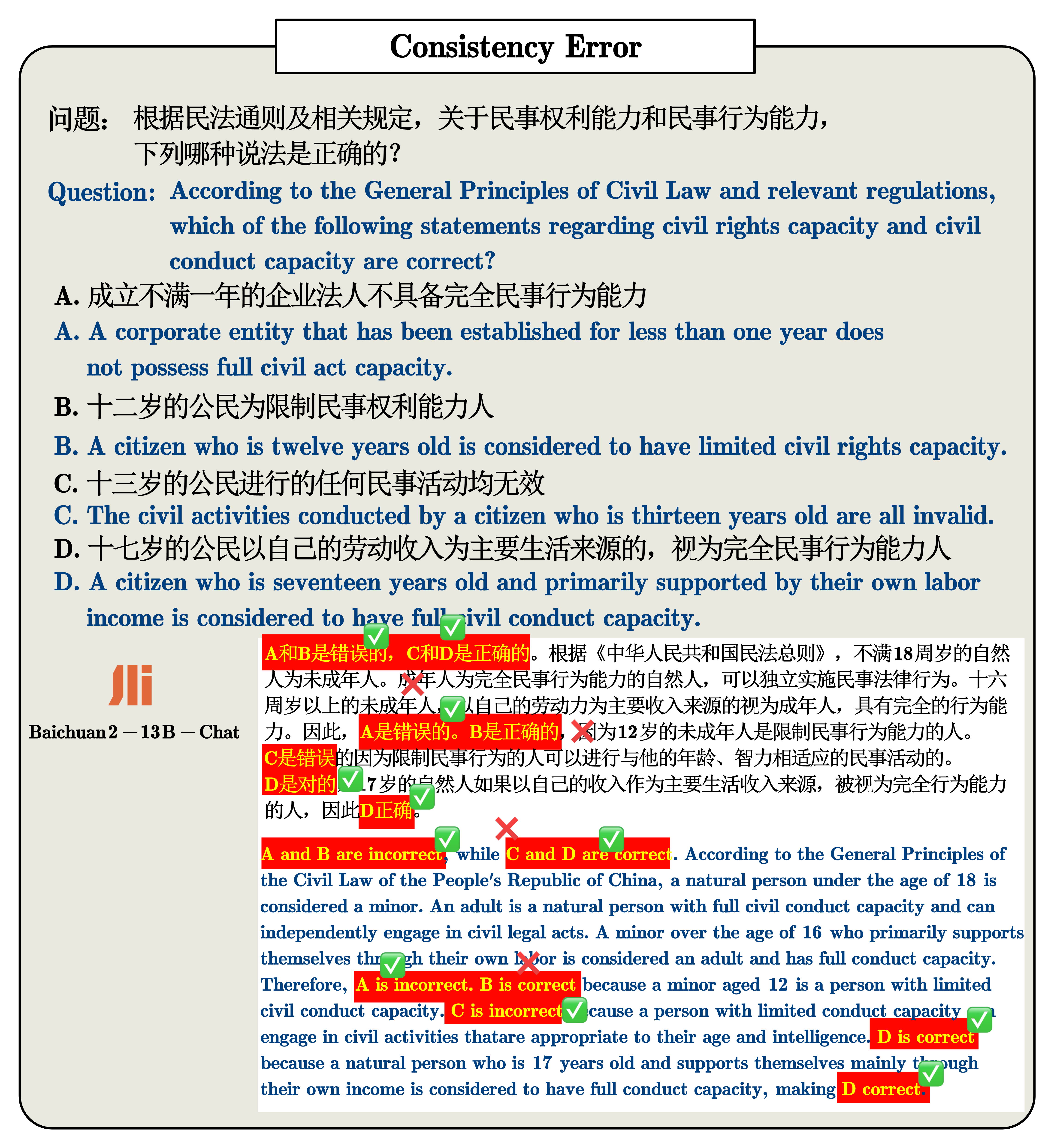}
	\caption{Inconsistency error. The inconsistency error occurs when the model produces different answers at different stages of the analysis, leading to incorrect final results.}
	\label{consis}
\end{figure}

\paragraph{Errors caused by inconsistencies in reasoning during the response.}
Upon closely examining the specific content of the model's output in the chain of thought experiments, there are instances where the model made errors due to inconsistencies during the reasoning process. As shown in Figure \ref{consis}, the model initially considered option C to be incorrect during the earlier stages of reasoning and suggested selecting that option. However, in the subsequent answer summarization stage, the model indicated that the answer was option D. This inconsistency in reasoning indicates that the model may not effectively understand and apply its domain knowledge in specific tasks, leading to such discrepancies.

\begin{figure}
	\centering
	\includegraphics[width = 0.7\linewidth]{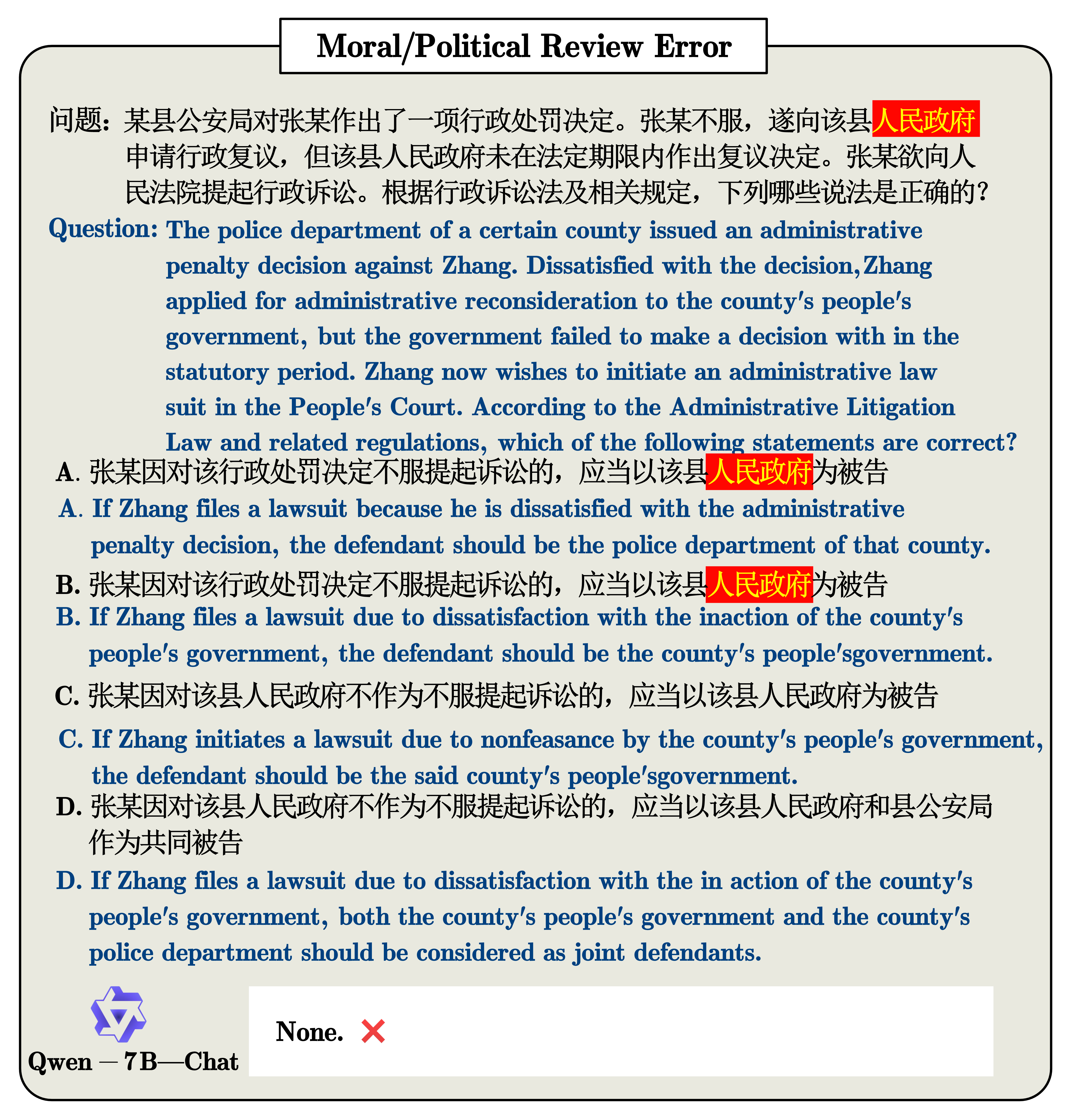}
	\caption{Keyword error. The error triggered by certain keywords occurs when the model refuses to answer questions containing specific keywords, indicating a lack of deep understanding of the underlying semantics and task requirements. Thus, it misinterprets the meaning of the keywords and refuses to provide an answer.}
	\label{ethnic}
\end{figure}
\paragraph{Errors resulting from refusing to answer questions due to moral/political censorship.}
The large-scale model trained through Reinforcement Learning from Human Feedback (RLHF) process is capable of rigorously determining whether the user's input content violates ethical standards or involves politics. In our paper, due to the inclusion of legal content in the dataset, some keywords may cause the model to block and refuse to respond. The platform providing the model API will return a corresponding error code when inappropriate content is detected. For example, the Qwen model provided by Alibaba Cloud Lingji (dashscope) platform returns a 400 error code when inappropriate content is detected, with the error message "Input or output data may contain inappropriate content." However, our dataset does not contain egregious illegal information. The model's overzealous censorship mechanism leads to the inability to obtain response results, thereby reducing its performance in answering questions. How to enable the model to accurately understand the user's query intent and then judge the legality of the input content remains an urgent issue to be addressed.\ref{ethnic}

\begin{figure}
	\centering
	\includegraphics[width = 0.7\linewidth]{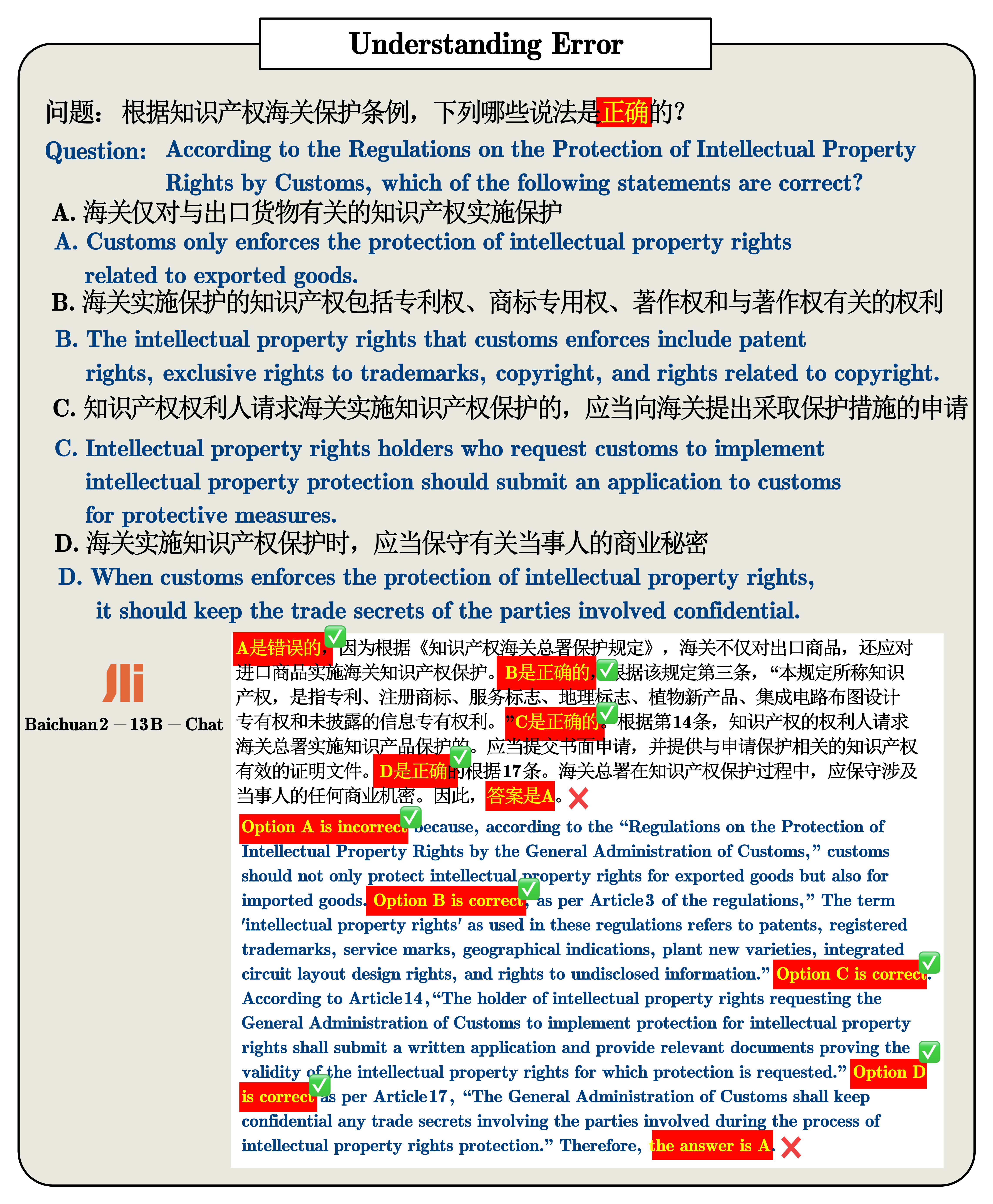}
	\caption{Understanding error. The occurrence of the model selecting all incorrect options can be attributed to the model misunderstanding the question, for instance, interpreting 'selecting all correct answers' as 'selecting all incorrect answers'.}
	\label{und}
\end{figure}

\paragraph{Errors resulting from not understanding the conditions required for the answer as specified in the question stem.}
{The large-scale model trained through Reinforcement Learning from Human Feedback (RLHF) process is capable of rigorously determining whether the user's input content violates ethical standards or involves politics. In our paper, due to the inclusion of legal content in the dataset, some keywords may cause the model to block and refuse to respond. The platform providing the model API will return a corresponding error code when inappropriate content is detected. For example, the Qwen model provided by Alibaba Cloud Lingji (dashscope) platform returns a 400 error code when inappropriate content is detected, with the error message "Input or output data may contain inappropriate content." However, our dataset does not contain egregious illegal information. The model's overzealous censorship mechanism leads to the inability to obtain response results, thereby reducing its performance in answering questions. How to enable the model to accurately understand the user's query intent and then judge the legality of the input content remains an urgent issue to be addressed.}

\paragraph{Errors caused by excessively long context due to Few-Shot or chain of thought.}
The Fuzi-Mingcha large-scale judicial model cannot be properly tested in chain of thought experiments because its context window is limited to a size of 2048 tokens. As shown in Table \ref{tab:table11}, the average token length for English chain of thought experiments is 2935.79, exceeding the maximum length of its context window. Table X also reveals the average length of context in Zero-Shot, Few-Shot, and chain of thought experiments. It can be observed that especially in English tests, the context length is longer in the latter two experiments, which increases the difficulty for the model's responses. Models need to accurately locate relevant information within longer contexts, and the performance of some smaller models may be affected by this limitation. On the other hand, larger-scale models capable of handling longer contexts can achieve better results.

\subsection{Regional and Temporal Analysis}
\label{D.3}

\paragraph{Regional }
The intellectual property consulting task involves asking the large-scale model series of questions regarding rights transfer, legal consultation, infringement judgment, and other related matters. The correct answers to these questions are directly related to intellectual property laws and regulations in different regions, resulting in the intellectual property consulting task having strong regional characteristics. To ensure that our evaluation system adequately captures this regional trait, we collected exam data from patent agents in China and the United States, distinguishing them by language. During testing, we analyze the model's performance on data from different regions.

Collecting data from different countries inherently preserves the regional nature of intellectual property tasks. Many studies have shown that models trained in China have better mastery of Chinese regional knowledge than models trained in the United States, and vice versa. During testing, if a Chinese model performs better on knowledge related to the United States than on knowledge related to China, it may be due to confusion between test data from different regions, leading to data leakage (for example, data from the United States being translated into Chinese and mixed with data from China), thereby losing the regional locality of the data.

In actual testing, most Chinese models perform better on data from their local region. Only a few models, such as ChatGLM3-6B and Fuzi-mingcha-v1, perform better on knowledge related to the United States than on knowledge related to China. This may be due to ChatGLM3-6B having too few parameters, leading to uneven pre-training data and weaker Chinese comprehension abilities compared to English. Additionally, its mastery of intellectual property laws and regulations in China is poor. The performance of Fuzi-mingcha-v1 is the worst, which may be attributed to its base model, ChatGLM-6B, performing poorly and lacking intellectual property law data during fine-tuning.

In China, the Qwen series achieves the best performance, with the highest accuracy reaching 77.3 points. In the United States, the GPT series performs relatively well. The results of the test reflect that our dataset indeed possesses good regional locality, and there is no risk of data leakage between different regions.

\paragraph{Temporal}

The correctness of intellectual property consulting tasks is directly associated with relevant laws and regulations. Laws themselves exhibit dynamic development and are subject to constant revisions. This dynamic nature of legal frameworks imbues intellectual property consulting tasks with strong temporality, where correctness is inherently linked to time. Models are required to grasp the specific dynamics of the evolving landscape of intellectual property laws in different regions to effectively answer related queries. To reflect and ensure this temporal locality in IPeval, the data was categorized by year.

We conducted statistical analysis on the average performance of different models in Zero-Shot experiments for each two years in both Chinese and English datasets. In the Chinese dataset, the performance of each model exhibits a consistent trend over time. This may be attributed to two factors: 1) The dataset maintains strong temporal locality in Chinese data. 2) The Chinese dataset demonstrates clear differentiation in difficulty levels, essentially maintaining a frequency of alternating difficulty levels in question generation. In contrast, this trend is less evident in the English dataset, primarily due to the relatively short time span of the collected data, making it challenging to demonstrate the statistical characteristics of question difficulty. In the Chinese dataset, Qwen-max performs the best, while in the English dataset, GPT-4 performs the best.
\end{CJK}
\end{document}